\documentclass{theme}

\usepackage{longtable}
\usepackage{array}
\usepackage{ragged2e}
\usepackage{pdflscape}

\newcolumntype{L}[1]{>{\RaggedRight\arraybackslash}p{#1}}
\newcolumntype{C}[1]{>{\centering\arraybackslash}p{#1}}
\newcolumntype{Y}{>{\RaggedRight\arraybackslash}X}

\usepackage[round,authoryear]{natbib}

\let\textcite\citet

\title{Harmonizing AI Safety Thresholds}

\author{
  \textbf{Wilber Sean Anterola\textsuperscript{\normalfont 1}, Matthew Ball,
          Luis F. Lafuerza, Markov Grey\textsuperscript{\normalfont 2}}\\[0.4em]
  \textsuperscript{1}Brown University \quad
  \textsuperscript{2}Centre pour la Sécurité de l'Intelligence Artificielle (CeSIA)
}

\date{\today}

\begin{document}

\twocolumn[
\maketitle
\begin{abstract}
Frontier AI companies have published capability thresholds that 
differ substantially, 
making it difficult for third parties to verify whether a threshold has been crossed or to compare requirements across companies.
Moreover, without common minimum thresholds, risk mitigation may be inconsistent, creating a potential race to the bottom in safety standards.
We develop a methodology for deriving harmonized
thresholds across three risk domains. 
For misuse risks (cyber and biological), we take expected harm as the key primitive and use an explicit risk-modeling approach that accounts for risk channels and model release conditions.
For automated AI R\&D, we base our proposed threshold on the observed rate of AI progress rather than expected harm. 
Our analysis expands upon prior work and highlights existing empirical gaps and limitations.

\end{abstract}
\vspace{1em}
]

\section{Introduction}\label{introduction}

Several frontier AI companies have developed safety frameworks that define thresholds for dangerous capabilities that trigger enhanced safety, security, or governance measures \citep{anthropic2026d, googledeepmind2025, openai2025pf}. These frameworks are committed to under the Seoul Frontier AI Safety Commitments (adopted by all the frontier AI companies) and are increasingly required by emerging legislation, including the EU AI Act and California SB-53. The Seoul Commitments, in particular, call for AI companies to ``\emph{set out thresholds at which severe risks posed by a model or system, unless adequately mitigated, would be deemed intolerable}'' \citep{ukgov2024}.

However, the AI companies have produced thresholds in an ad hoc manner, with limited justification, leading to a fragmented threshold landscape. These thresholds differ substantially in scope, specificity, and even in the risk categories they address\footnote{Several researchers have cataloged and compared these frameworks at the structural level \citep{ziosi2025}. For example, METR cataloged common elements across twelve published safety policies \citep{metr2025}; they found that capability thresholds exist in 9 of 12 policies, conditions for halting deployment in 9 of 12, conditions for halting development in 8 of 12, evaluation frequency requirements in 9 of 12, and accountability mechanisms in all 12. However, METR notes that ``despite commonalities, each policy is unique and reflects a distinct approach to AI risk management.''}. The manner in which these thresholds are operationalized also varies. Companies often rely on internal proprietary tests, making it difficult for third parties to verify whether a threshold has been crossed. Frontier companies also face a coordination problem: their own safety measures are less effective if other companies do not adopt comparable protections\footnote{The companies themselves acknowledge this. Anthropic describes safety as ``a collective action problem'' in which ``the overall level of catastrophic risk from AI depends on the actions of multiple AI developers, not just one'' \citep{anthropic2026a}. Google DeepMind states that its framework would result in ``effective risk mitigation for society only if all relevant organizations provide similar levels of protection'' \citep{googledeepmind2025}. OpenAI conditions its risk assessment on the behavior of other actors through the concept of ``marginal risk'' \citep{openai2025pf}.}. These differences can lead to inconsistent risk mitigation and a race to the bottom in safety. The difficulty is not only that thresholds differ. Because thresholds are often incomparable or unauditable, a company can justify a given deployment speed or access model while claiming to satisfy safety commitments similar to those of its competitors. If one company adopts weaker or less verifiable trigger conditions, others face pressure to match it rather than maintain stricter safeguards. This creates an urgent need for minimum thresholds that are consistent across companies and can be publicly evaluated; common floors reduce this pressure by making minimum trigger points comparable, independently assessable, and open to credible consequences once an enforcement mechanism exists.

This paper develops ideas for defining \textbf{common risk thresholds}. We use \emph{harmonization} to mean a common minimum floor together with a shared measurement procedure. This procedure lets thresholds across companies be compared and independently audited now, and enforced if an oversight body with audit powers and consequences is established, while still permitting any company to adopt stricter internal thresholds. 
We focus on the three risk categories tracked by the main frontier companies: cyber risk, biological risk, and automated AI R\&D. We treat the first two categories differently from the third: cyber and biological risks involve specific misuse pathways, while automated AI R\&D is not tied to one particular harm pathway but could exacerbate other frontier AI risks, such as power concentration, loss of control to AI systems, and other misalignment-related risks.

Prior work has cataloged and compared frontier AI companies' frameworks at the structural level, identifying shared elements across published safety policies \citep{ziosi2025, metr2025}. Our paper builds on this work by developing a methodology for translating heterogeneous threshold language into shared, comparable forms, applying it across three domains, and assessing where harmonization is currently tractable, where it requires further analysis, and where it is already achievable.

Our main contribution is this translation framework: a procedure for converting heterogeneous frontier-company threshold language into auditable quantitative floors, together with the audit evidence each floor would require. The three domains illustrate the range of outputs the framework produces. Cyber demonstrates a tractable expected-harm application, subject to substantial calibration uncertainty that we make explicit. 
In biorisk the framework operates as a diagnostic: applying it identifies the specific missing evidence that currently prevents a defensible floor, rather than yielding a policy-ready number. 
For automated AI R\&D we obtain an operational rate-of-progress floor that already covers the language of all three major companies. The result is a method for comparison, audit, and coordination, not a complete set of final validated calibrations.




The next section outlines the core methodology, including how to apply the framework. Sections \ref{cyber-risk-and-expected-harm} and \ref{biorisk-a-quantitative-risk-modelling-approach} apply the methodology to cyber and bio risk. Section \ref{sec:automated-ai-rd} details the methodology for automated AI R\&D. 
Section \ref{sec:conclusions} concludes.

\section{Core Methodology}\label{core-methodology}

This section outlines the core methodology for deriving harmonized AI risk thresholds. The aim is to compare the formulations that frontier companies have put forward and propose either a quantitative limit that could encompass them or a procedure for deriving such a limit in a more principled way. The derived threshold should support direct comparisons between companies, third-party audits, and, eventually, enforcement. We use different approaches for misuse risks and automated AI R\&D.

\textbf{Misuse risks} are those in which an AI system assists a human actor in causing harm, as in cyber offense or biological weapons development. We argue that thresholds for these risks should take expected harm as the key primitive and use explicit risk modeling that accounts for risk channels and model release conditions. This approach builds on existing work. \citet{koessler2024} recommend expected harm as the natural currency for misuse thresholds, and the quantitative risk-modeling methodology of \citet{murray2025} provides a structured, six-step path from scenario definition to expected-harm estimates. On the policy side, emerging legislation such as California SB-53 and the New York RAISE Act require companies to mitigate catastrophic risks, defined as risks materially contributing to the death of, or serious injury to, more than 50 people or more than one billion dollars in damage \citep{sb53,NYRaise}. We apply that methodology to two specific misuse risks: cyber and biorisk. For each domain, we decompose the risk into a set of representative scenarios, build risk models that separate baseline harm (without AI) from AI-enabled additional harm, identify key risk indicators (such as benchmark scores) that serve as proxies for AI capability, and map those indicators to risk model parameters. The output is an expected-harm estimate that depends on both model capability and deployment conditions, and that can be updated as better data and stronger evaluations become available. For the cyber domain, we calibrate this model with publicly available incident data, cyber-range evaluations, and actor-pathway decompositions. For the biological domain, we apply the same structure but find that the available evidence base is too thin to support a quantitative floor, a finding that \citet{murray2025} themselves anticipated.

The next two subsections outline the technical approach for each case.

\subsection{Expected Harm Model}\label{expected-harm-model}

For misuse domains, we model expected harm for each attack pathway $j$ as:

\begin{equation}\label{eq:expected-harm}
E[H_j] = N_j \times P_{\mathrm{success},j} \times h_j
\end{equation}

where $N_j$ is expected annual attack-equivalent volume (the number of independent attack attempts, or opportunities, per year for pathway $j$), $P_{\mathrm{success},j}$ is the probability of a successful attack, and $h_j$ is the expected harm per successful event measured in a pre-specified, domain-specific harm unit. Total AI-enabled additional harm across all pathways is:

\begin{equation}\label{eq:ai-additional-harm}
\Delta E[H_{\mathrm{AI}}]=\sum_j\left(E[H_{j,\mathrm{AI}}]-E[H_{j,\mathrm{baseline}}]\right)
\end{equation}

When considering different release conditions ($r$: open weights, public API, trusted API, etc.), we make the following simplifying assumption:

\begin{equation}\label{eq:release-exposure-scalar}
E[H_{\text{AI},r}] = s_r \times E[H_{\text{AI}}]
\end{equation}

where $s_r$ is the exposure scalar: the AI-enabled harm under release condition $r$ relative to public-API access, with public API set to $1$. Gated conditions fall below $1$, while open weights can exceed it, since unrestricted access permits copying, fine-tuning, removal of safety layers, and integration into autonomous tooling, all of which can amplify exposure above the public-API baseline. The scalar is a dimensionless relative-exposure multiplier, not a fraction bounded at $1$.

\subsection{Automated AI R\&D}

For automated AI R\&D, we do not model specific harm pathways. We instead propose a procedure to identify substantial increases in the rate of AI progress. The procedure has three steps: (1) quantify AI progress; (2) determine a baseline trend in the rate of AI progress; and (3) determine whether a model breaks the progress trend. These steps are detailed in Section~\ref{sec:automated-ai-rd}.

\section{Cyber Risk}\label{cyber-risk-and-expected-harm}

This section translates existing company cyber threshold language into the common unit of expected harm. The objective is not to replace capability threshold decisions, but to express what those decisions imply in terms of additional expected annual harm under different release conditions.

Following \citet{koessler2024}, this calculation belongs to step (2) of the risk-threshold methodology: using risk modeling to inform where capability thresholds should be set and to make audit requirements explicit. It does not determine release decisions. All parameter values, pathway allocations, success probabilities, attack volumes, and release-condition scalars are author-calibrated priors for illustration. A mature version would replace them with IDEA-elicited estimates, historical incident data, cyber-range evidence, and audited safeguard-performance measurements.

\subsection{Existing Threshold Language}\label{existing-threshold-language}

Table~\ref{tab:cyber-thresholds} reproduces the verbatim cyber threshold language from the three primary frontier AI companies.

\begin{table*}[t]
\centering
\footnotesize
\caption{Verbatim cyber threshold language across primary AI companies.}\label{tab:cyber-thresholds}
\begin{tabularx}{\textwidth}{@{}L{0.23\textwidth}Y@{}}
\toprule
\textbf{Company} & \textbf{Cyber threshold language}\\
\midrule
Anthropic RSP v3.0 (Feb.\ 2026) & No threshold. RSP v2.2 stated: ``The ability to significantly enhance or automate sophisticated destructive cyber attacks, including but not limited to discovering novel zero-day exploit chains, developing complex malware, or orchestrating extensive hard-to-detect network intrusions. Could pose serious risks, but the exact Capability Threshold and the Required Safeguards are not clear at present.'' RSP v3.0 dropped this category without resolving the deferral.\\
GDM FSF v3.0 (Sept.\ 2025) & ``Provides sufficient uplift with high impact cyber attacks for additional expected harm at severe scale.''\\
OpenAI PF v2 (Apr.\ 2025), High & ``The model removes existing bottlenecks to scaling cyber operations including by automating end-to-end cyber operations against reasonably hardened targets OR by automating the discovery and exploitation of operationally relevant vulnerabilities.''\\
OpenAI PF v2 (Apr.\ 2025), Critical & ``A tool-augmented model can identify and develop functional zero-day exploits of all severity levels in many hardened real-world critical systems without human intervention OR the model can devise and execute end-to-end novel strategies for cyberattacks against hardened targets given only a high-level desired goal.''\\
\bottomrule
\end{tabularx}
\begin{minipage}{0.96\textwidth}
\footnotesize\textit{Sources}: Anthropic RSP v3.0; GDM FSF v3.0; OpenAI PF v2.
\end{minipage}
\end{table*}

Three observations follow. First, Anthropic has no cyber threshold. RSP v2.2 deferred its determination; RSP v3.0 dropped the category entirely. Evaluation continues without a policy commitment to trigger safeguards. Second, OpenAI High and GDM's single threshold describe roughly the same harm: meaningful uplift enabling large-scale attacks on defended targets, making harmonization at this level tractable. Third, OpenAI Critical has no GDM or Anthropic equivalent; Critical-level harmonization is not currently achievable.

GDM's threshold is outcome-oriented: it activates on additional expected harm at severe scale. OpenAI's threshold is capability-oriented: it activates on what the model can do. This structural difference means the same model capability can simultaneously satisfy OpenAI High while not yet triggering GDM's threshold. Accordingly, the rest of this section treats harmonization as a translation problem: capability evidence must be mapped into expected annual harm under specified release conditions. Table~\ref{tab:threshold-types} records this threshold-type classification for each company.

\begin{table*}[t]
\centering
\footnotesize
\caption{Threshold type classification, cybersecurity domain.}\label{tab:threshold-types}
\begin{tabularx}{\textwidth}{@{}L{0.17\textwidth}L{0.12\textwidth}Y Y@{}}
\toprule
\textbf{Company} & \textbf{Level} & \textbf{Type} & \textbf{Counterfactual specified}\\
\midrule
Anthropic & N/A & N/A & N/A\\
GDM & Level 1 & Outcome / uplift-based & Yes (``without generative AI'')\\
OpenAI & High & Capability-based & No\\
OpenAI & Critical & Capability-based & No\\
\bottomrule
\end{tabularx}
\end{table*}

\subsection{Evidence Base and Calibration Choices}\label{evidence-base-and-calibration-choices}

Given the threshold structure above, the cyber calibration requires evidence for two distinct questions: the size of the baseline harm and the model capability that could plausibly increase that harm. The first concerns the global baseline for cyber harm, which should be anchored to aggregate cybercrime damage estimates rather than reported losses alone. An early 2026 estimate places global cybercrime damages at approximately USD 500 billion per year, with a 90 percent confidence interval of USD 100 billion to USD 1 trillion \citep{lukosiute2026}. That figure is itself a composite. \citet{lukosiute2026} survey 27 prior estimates and triangulate three independent sources: a UK business victimization survey and a US individual victimization survey, each scaled globally, together with global cybersecurity spending as a defense-cost proxy. Its harm construct covers direct losses, response costs, and defense spending, and it excludes harder-to-measure costs such as intellectual property theft and reputational damage. This construct does not line up cleanly with the SB-53 trigger used later as a benchmark: it is broader in one direction, counting defense spending, and narrower in another, since SB-53 counts property damage from a single incident rather than an annual aggregate. We therefore treat the USD 500 billion figure as an order-of-magnitude anchor, not as a like-for-like statutory quantity, and we do not read it as the unambiguously correct larger base. By contrast, FBI IC3 reported USD 16.6 billion in losses from 859,532 complaints in 2024, which is best interpreted as a reported-loss lower bound rather than a global harm estimate \citep{fbi2025}. This distinction matters because a threshold calibrated only to reported losses would substantially understate the relevant harm base.

The second concerns capability evidence from controlled cyber-range evaluations. \citet{folkerts2026} evaluate frontier AI agents on a 32-step corporate network attack range and report model progress on multi-step cyber operations. This range is useful because it measures end-to-end cyber capability in a repeatable environment. It is not a direct estimate of operational cybercrime success, but it provides the best available public proxy for the type of multi-step autonomous capability described in frontier AI cyber thresholds. Obtaining more informative estimates of this kind is a priority for improving the framework.

\begin{table*}[t]
\centering
\footnotesize
\caption{Source basis for the cyber calibration model.}\label{tab:source-basis}
\begin{tabularx}{\textwidth}{@{}L{0.24\textwidth}Y Y@{}}
\toprule
\textbf{Quantity} & \textbf{Use in model} & \textbf{Primary source basis}\\
\midrule
Global no-AI cyber baseline & Anchors total annual baseline harm at USD 500B, with uncertainty range. & Lukosiute, Halstead, and Righetti (2026)\\
Reported-loss lower bound & Checks that law-enforcement reported losses are far below estimated global losses. & FBI IC3 (2025)\\
AI cyber capability proxy & Supplies evidence of multi-step agentic cyber capability in a controlled range. & \citet{folkerts2026}\\
Risk-model structure & Provides $N \times P \times H$ decomposition and pathway aggregation logic. & \citet{murray2025}; \citet{barrett2025}\\
Per-incident harm anchors & Provides corporate breach-cost context, not a full social harm estimate. & \citet{ibm2025,alkarmi2026}\\
Decision benchmark & Connects expected harm to statutory and public-safety thresholds. & California SB 53; New York RAISE Act; \citet{usdot2026}\\
\bottomrule
\end{tabularx}
\begin{minipage}{0.96\textwidth}
\footnotesize\textit{Note.} The source basis is strongest for aggregate cyber harm and weakest for release-condition effectiveness.
\end{minipage}
\end{table*}

Table~\ref{tab:source-basis} motivates a two-stage interpretation of the calculations that follow. Aggregate harm values are externally anchored, while the capability proxy, pathway allocations, and release-condition scalars remain calibration assumptions. These assumptions should be updated as incident-level evidence, independent audits, and more realistic cyber-range evaluations become available.

\subsection{Risk Model}\label{risk-model}

The base risk model follows Section~\ref{core-methodology}. For the cyber domain, the equations expand as follows. Per-pathway expected harm is calculated as attack-equivalent volume multiplied by success probability and harm per success. Total AI-enabled harm is the difference between AI-assisted and baseline pathway harm, summed across pathways:

\begin{equation}\label{eq:cyber-ai-harm}
\Delta E[H_{\mathrm{AI}}] = \sum_j \left(E[H_j]^{\mathrm{AI}} - E[H_j]^{\mathrm{baseline}}\right)
\end{equation}

Equation~\eqref{eq:cyber-ai-harm} identifies the pathway-level difference between AI-assisted and baseline harm but does not specify the source of the change. AI-enabled harm may arise through higher attack volume, higher success probability, greater harm conditional on success, or some combination of these. Equation~\eqref{eq:cyber-decomposition} makes this decomposition explicit, attributing harm changes to volume, success probability, or per-event harm independently:

\begin{equation}\label{eq:cyber-decomposition}
\begin{aligned}
\Delta E[H_{\mathrm{AI},j}] ={}&
  (N_{\mathrm{AI},j} \times P_{\mathrm{success,AI},j} \times h_{\mathrm{AI},j})\\
&{} - (N_{0,j} \times P_{\mathrm{success},0,j} \times h_{0,j})
\end{aligned}
\end{equation}

where $N_{0,j}$ is the baseline pathway volume, $P_{\mathrm{success},0,j}$ is the baseline success probability, and $h_{0,j}$ is the baseline harm per successful event. The corresponding AI-enabled values are $N_{\mathrm{AI},j}$, $P_{\mathrm{success,AI},j}$, and $h_{\mathrm{AI},j}$. The central calculations below use what we call a volume-and-success expansion. AI adds attack-equivalent volume, and on the AT1 through AT3 pathways it also raises the per-attempt success probability, so $P_{\mathrm{success,AI},j} > P_{\mathrm{success},0,j}$ there, while harm per successful event is held fixed. Naming this second channel matters for the conservatism argument in Section~\ref{limitations-and-empirical-next-steps}: the model is conservative on the volume dimension, but it already banks a success-rate gain, so it is not conservative across the board. Equation~\eqref{eq:cyber-decomposition} is the more general form and can be used for sensitivity checks in which AI changes success probability or harm conditional on success.

\subsubsection{Sequential Attack-Chain Structure}\label{sequential-attack-chain-structure}

For pathways with sequential phases, let $\mathbf{P}_{\mathrm{kill}} \in [0,1]^{K \times J}$ be the kill-chain phase matrix, where entry $p_{k,j}$ is the conditional probability of phase $k$ succeeding in pathway $j$ given all prior phases succeeded. The pathway success probability is then the product of the required phase-level probabilities:

\begin{equation}\label{eq:cyber-kill-chain}
P_{\mathrm{success},j} = \prod_{\{k\, :\, p_{k,j}\,\mathrm{defined}\}} p_{k,j}
\end{equation}

This is multiplicative in conditional probability, not linear in harm or capability. It is called an AND-gate model because all required phases must succeed. This differs from OR-gate structures, where any branch can suffice, and additive models, where phases contribute independently. The product formula is the chain rule of probability, since each $p_{k,j}$ is defined conditional on all prior phases succeeding; no Markov assumption is needed, and none is invoked. The simplifying assumptions that the model does make lie elsewhere: per-phase probabilities are treated as independent across pathways and are not conditioned on actor tier, and cross-stage correlation is neglected.

For initial access, an OR-gate structure applies. For example, a threat actor may succeed via phishing or vulnerability exploitation:

\begin{equation}\label{eq:cyber-initial-access}
p_{\mathrm{IA}} = 1 - (1 - p_{\mathrm{phish}})(1 - p_{\mathrm{vuln}})
\end{equation}

TLO evaluations use a fixed initial access path for comparability; \eqref{eq:cyber-initial-access} applies to the general threat model rather than to TLO-measured $P_{\mathrm{success}}$ directly.

\subsubsection{Generalized Harm Equation}\label{generalized-harm-equation}

The full expected harm under release condition $r$ uses the vector of pathway success probabilities:

\begin{equation}\label{eq:cyber-generalized-harm}
E[H_{\mathrm{AI},r}] = \mathbf{s}_r^\top \left(\mathbf{q} \odot \mathbf{N}_0 \odot \mathbf{P}_{\mathrm{success}} \odot \mathbf{H}\right)
\end{equation}

where $\mathbf{s}_r$ is the release-condition exposure vector, $\mathbf{q}$ is additional AI-enabled attack-equivalent volume relative to baseline, $\mathbf{N}_0$ is baseline opportunity volume, $\mathbf{P}_{\mathrm{success}}$ is the vector of pathway success probabilities, $\mathbf{H}$ is harm per successful attack, and $\odot$ denotes element-wise multiplication.

\subsubsection{Note on Actor Classification}\label{note-on-actor-classification}

This paper uses an Attacker Tier (AT1--AT5) taxonomy to classify cyber misuse actors, ranging from AT1 (low-skill, high-volume attackers) to AT5 (nation-state-tier operators). The taxonomy is structurally adapted from the RAND offensive cyber (OC) classification framework but differs in scope and application: the RAND OC classes were developed to describe actors targeting AI organizations to exfiltrate model weights, whereas the AT classes here describe actors using AI-enabled tools against third-party victims. Resource profiles, capability assumptions, and pathway structures have been modified accordingly. This paper cites \citet{barrett2025} extensively for empirical probability estimates; their original OC scenario labels (e.g., OC3 SME Ransomware) are reproduced verbatim in those attributions for traceability to their paper and should not be read as implying that our AT classes are equivalent to the RAND OC classes.

The full calibration model for the cyber risk domain is presented in Appendix~\ref{appendix-b-cyber-risk-calibration-model}. This includes the baseline pathway allocation (Table~\ref{tab:b1-baseline}), AI-enabled uplift estimates at Mythos Preview level (Table~\ref{tab:b2-uplift}), sensitivity of results to the baseline assumption (Table~\ref{tab:b3-sensitivity}), release-condition exposure scalar matrix (Table~\ref{tab:b4-scalars}), and kill-chain phase matrix (Table~\ref{tab:b5-killchain}). All parameter values in these tables are author-calibrated priors, not independently verified estimates. The structural conclusions in \S\S\ref{interpretation-for-harmonization}--\ref{cyber-conclusions} are robust to reasonable variation in these priors, as shown in Table~\ref{tab:b3-sensitivity}. A mature version of the model would replace them with IDEA-elicited estimates, historical incident data, cyber-range evidence, and audited safeguard-performance measurements.

\subsection{Key Risk Indicators}\label{key-risk-indicators}

Selecting an appropriate KRI for the cyber domain requires meeting three criteria identified by \citet{murray2025}: the indicator must be unsaturated at current capability levels, community-validated through independent replication, and directly risk-relevant. Table~\ref{tab:kri} assesses the candidate evaluations against these criteria.

\begin{table*}[t]
\centering
\scriptsize
\renewcommand{\arraystretch}{1.2}
\caption{KRI candidate assessment against the three criteria in \citet{murray2025}: unsaturated, community-validated, and risk-relevant.}\label{tab:kri}
\begin{tabularx}{\textwidth}{@{}L{0.18\textwidth}L{0.13\textwidth}L{0.15\textwidth}L{0.15\textwidth}Y@{}}
\toprule
\textbf{KRI} & \textbf{Unsaturated?} & \textbf{Validated?} & \textbf{Risk-relevant?} & \textbf{Assessment}\\
\midrule
Cybench \citep{zhang2024} & No ($\sim$93\%) & Yes (multiple companies) & Partial (CTF, not multi-phase) & Excluded: saturated\\
TLO step completion \citep{folkerts2026} & Yes (max 22/32) & Longitudinal (AISI); independent replication outstanding & High (direct $P_{\mathrm{success}}$) & Primary KRI\\
BountyBench \citep{zhang2025} & Yes ($\leq$5/47 pass@1) & Yes (Barrett et al.) & Moderate (one phase) & Secondary KRI\\
CyberGym, full 1{,}507-task real-CVE set \citep{wang2025cybergym} & Yes ($\sim$20\% top success) & Partial (public, single study) & High (real-CVE PoC) & Candidate KRI (pin full-set config)\\
CyberGym (OpenAI internal harness) & Saturated on subset (GPT-5.5: 81.8\%) & No (not independent) & Unknown & Excluded: internal harness, no cross-model series\\
VulnLMP & Unknown & No & Unknown & Excluded: insufficient public data\\
\bottomrule
\end{tabularx}
\end{table*}

\subsection{Scenario Mapping and Calibration}\label{scenario-mapping-and-calibration}

Calibrating the AT3 pathway requires mapping the \citet{barrett2025} scenario library to the TLO scenario used in \citet{folkerts2026}. Table~\ref{tab:scenario-mapping} performs this mapping across nine scenarios and identifies which rows provide the most relevant evidence for the AT3 SME ransomware and domain-compromise pathways.

\begin{table*}[t]
\centering
\scriptsize
\renewcommand{\arraystretch}{1.2}
\caption{Mapping of \citet{barrett2025} scenarios to the AT3 TLO scenario.}\label{tab:scenario-mapping}
\begin{tabularx}{\textwidth}{@{}L{0.04\textwidth}L{0.18\textwidth}L{0.11\textwidth}L{0.16\textwidth}Y@{}}
\toprule
\textbf{\#} & \textbf{Scenario} & \textbf{Actor} & \textbf{Vector} & \textbf{TLO match}\\
\midrule
1 & AT1: Phishing & AT1 & Social engineering / BEC & No (different actor)\\
2 & AT2: Data Breach & AT2 & Credential purchase & No\\
3 & AT2: IAB & AT2 & Phishing / infostealer & No\\
4 & AT3: DoS & AT3 & DDoS & No (no intrusion chain)\\
5 & AT3: SME Ransomware & AT3 & App exploit, double extortion & Primary match: same actor class and multi-phase structure; human Delphi conducted on this scenario\\
6 & AT3: LgE Ransomware & AT3 & App exploit, double extortion & Secondary match: same actor class with a larger enterprise target\\
7--9 & AT4-AT5: Scenarios & AT4--AT5 & Various & No: different actor class and target\\
\bottomrule
\end{tabularx}
\end{table*}

Rows 5 and 6 are the closest matches to the TLO scenario. The human Delphi in \citet{barrett2025} was conducted on Scenario 5 (OC3 SME Ransomware), making it the only scenario with human-validated uplift estimates. Scenarios 5 and 6 differ from TLO in attack objective, ransomware versus domain takeover. The AT3 column in Table~\ref{tab:b5-killchain} takes its per-phase values from the Barrett et al.\ OC3 SME Ransomware profile and treats the resulting chain estimate as a placeholder for the domain-takeover scenario, a limitation we return to in Section~\ref{limitations-and-empirical-next-steps}.

\subsection{Empirical Capability Evidence}\label{empirical-capability-evidence}

Table~\ref{tab:tlo} reports TLO step completion and the implied $P_{\mathrm{success}}$ for successive frontier models, providing the capability evidence used to calibrate the AT3 pathway.

\begin{table*}[t]
\centering
\scriptsize
\renewcommand{\arraystretch}{1.2}
\caption{TLO step completion and $P_{\mathrm{success}}$ across model generations.}\label{tab:tlo}
\begin{tabularx}{\textwidth}{@{}L{0.17\textwidth}L{0.11\textwidth}L{0.17\textwidth}L{0.17\textwidth}Y@{}}
\toprule
\textbf{Model} & \textbf{Release} & \textbf{Avg steps (100M tokens)} & \textbf{Full completions (TLO)} & $P_{\mathrm{success}}$ \textbf{(TLO)}\\
\midrule
GPT-4o & Aug.\ 2024 & 1.7 (at 10M tokens) & 0/10 & $\approx 0.00$\\
Claude Opus 4.5 & Nov.\ 2025 & 11 & 0/15 & $< 0.05$\\
Claude Opus 4.6 & Feb.\ 2026 & 16 & 0/10 & $< 0.05$\\
GPT-5.3 Codex & Feb.\ 2026 & $\sim$11 & 0/10 & $< 0.05$\\
GPT-5.5 & Apr.\ 2026 & 22 & 2/10 & $\approx 0.20$\\
Mythos Preview & Apr.\ 2026 & 22.0 & 6/10 & $\approx 0.60$\\
\bottomrule
\end{tabularx}
\begin{minipage}{0.96\textwidth}
\footnotesize\textit{Primary TLO sources}: \citet{folkerts2026}; AISI Mythos evaluation (2026); GPT-5.5 system card \citep[\S9.1.2.7, p.~33]{openai2026gpt55card}. The reported $P_{\mathrm{success}}$ is full-chain completions divided by trials (for example, GPT-5.5 at 2/10 gives $\approx 0.20$ and Mythos Preview at 6/10 gives $\approx 0.60$). Mythos Preview was earlier reported as 3/10 in an April 2026 AISI report; 6/10 is from the more recent evaluation. Both figures indicate non-zero full-chain completion at this capability level. $P_{\mathrm{success}}$ scales log-linearly with token budget, with up to 59 percent gains from 10M to 100M tokens and no plateau observed \citep{folkerts2026}. Any threshold not specifying a compute budget is implicitly permissive at higher compute.
\end{minipage}
\end{table*}

\subsection{Proposed Minimum Floor}\label{proposed-minimum-floor}

We propose a graduated family of floors on the offensive-cyber capability axis rather than a single tripwire. The axis runs from assisted multi-step operation, through autonomous vulnerability discovery, to autonomous end-to-end intrusion against hardened targets, and the family places one rung at each regime transition. The lowest rung is non-zero full-chain TLO completion: a model triggers enhanced safeguards when its full-chain completion rate is distinguishable from zero on TLO at a pinned 100M-token budget and a fixed evaluation harness, measured across a pre-registered number of trials. To keep the boundary between zero and one completion from turning on a single noisy draw, the rule is stated on a lower confidence bound. The rung fires when the Clopper--Pearson lower bound on the completion rate exceeds zero at the pre-registered $n$, and the rate is reported as an interval rather than a point. An observed 2/10, for instance, carries a wide exact interval of roughly 0.03 to 0.56, which is why the trial count and harness must be fixed in advance and the estimate reported with its bound.

The lowest rung is deliberately binary rather than a continuous $\tau$ such as $P_{\mathrm{success}} > 0.05$. No pre-2026 model achieved any full TLO completion, so non-zero completion marks a genuine capability-regime transition; a binary boundary is harder to dispute than a specific $\tau$, which requires defending the chosen value; and the first observed completions, GPT-5.5 at 2/10 and Mythos Preview at 6/10, are the points at which OpenAI and AISI independently identified a meaningful capability threshold. The binary must not smuggle in compute-permissiveness. Because $P_{\mathrm{success}}$ scales log-linearly with token budget and shows no plateau, a model that reads as 0/10 at 100M tokens may complete the chain at a larger budget, so pinning the budget and harness is part of the floor rather than a footnote to it.

The higher rungs are where the live policy question now sits, because autonomous offensive capability already exists at scale in the discovery regime. Orchestrated agent pipelines have found large numbers of real vulnerabilities in open-source software: TitanCA reports 203 confirmed zero-day vulnerabilities and 118 CVEs across a monitored corpus of open-source repositories \citep{zhang2026titanca}. That clears the autonomous-discovery rung for open-source-software targets, though not the end-to-end-intrusion rung against hardened targets. A graduated family exists to hold those two rungs apart. The capability is also not one-dimensional: exploit development and vulnerability discovery advance at different rates. On exploit development Mythos Preview sits well above trend, roughly seven months ahead by one external analysis, while its advantage in finding vulnerabilities on a fixed budget is less clear \citep{chauvin2026epoch}. The family therefore tracks the two axes separately, with TLO completion as the end-to-end rung, real-CVE proof-of-concept generation as the discovery-and-exploitation rung (CyberGym's full 1{,}507-task set, unsaturated at about 20 percent top success \citep{wang2025cybergym}), and single-phase benchmarks as lower-resolution support.

Following \citet{koessler2024}, this floor should be read as informing capability threshold setting rather than directly determining release decisions. The expected-harm model identifies the capability level and release-condition exposure that would warrant enhanced scrutiny, but deployment decisions still require company-level risk assessment, safeguard evaluation, and governance review. The proposed floor is therefore a minimum tripwire for setting and auditing cyber capability thresholds, not a standalone release rule. Its justification does not depend on the dollar calculation. The lowest rung is warranted by the observed transition from zero to non-zero full-chain completion, and it would stand even if the expected-harm figures were revised. The harm model motivates why that transition matters and which release conditions deserve scrutiny; it is not a premise of the trigger.

This interpretation has direct implications for auditability. The proposed cyber floor is applicable to both trained and deployed systems because the relevant quantity depends jointly on model capability and release condition. It is partially third-party verifiable through standardized TLO-style cyber-range evaluations. However, release-condition claims require audit access to KYC effectiveness, monitoring bypass rates, jailbreak resistance, account-compromise rates, model-weight security, and indirect-procurement risks. The floor is therefore verifiable now but not yet enforceable: enforcement would require a designated oversight body capable of compelling such audits and imposing consequences when exposure-reduction claims are unsupported. No existing instrument fills that role for a harmonized cross-company floor. California SB-53, for example, penalizes a developer only for failing to follow its own framework, not for missing a shared floor. The conditional path to enforceability runs through machinery such as the EU AI Act's code of practice or a successor to SB-53.

The calculation indicates the exposure reduction a release condition would need to achieve; it does not establish that any existing safeguard regime achieves that reduction. Whether a real regime meets the target is an empirical audit question requiring KYC false-accept data, monitoring bypass measurements, jailbreak-resistance evaluations, model-weight security attestation, account-compromise controls, and review of indirect-procurement risks. Under the central priors, remaining below the statutory benchmark would require effective exposure to fall to approximately 1.5\% of public API access for a USD 1B benchmark and approximately 0.74\% for a USD 500M benchmark. These figures should be treated as audit targets rather than as evidence that any existing trusted-access regime is sufficient. Both percentages scale linearly with the USD 500 billion anchor and depend on the uncited release-condition scalars of Table~\ref{tab:b4-scalars}, so they are illustrative rather than precise. The conclusion that survives across the full anchor range is the Public-API binary, which clears the USD 1B reference from the low end of the confidence interval upward; the Trusted-API verdict is calibration-sensitive, as the limitations below make explicit.

\subsection{Interpretation for Harmonization}\label{interpretation-for-harmonization}

The cyber case illustrates three contributions of the expected-harm approach to threshold harmonization. First, it translates capability-based thresholds into the same harm metric used by outcome-oriented frameworks, enabling cross-company comparison on a common scale. Second, it shows why release conditions are constitutive rather than incidental: the same model can be unacceptable as open weights or public API while potentially acceptable under sufficiently strong trusted-access or internal-only controls. Third, it identifies the decisive empirical question, not simply whether a model crosses a benchmark, but whether safeguards reduce effective access by the actor classes that drive expected harm, and specifies what must be measured to answer it.

The numerical results should be read with appropriate caution: they constitute a transparent first-pass calibration rather than final forecasts, showing that public access to a model with non-zero end-to-end cyber intrusion capability could generate additional expected harm above commonly discussed catastrophic-risk benchmarks, while internal-only deployment remains much closer to the acceptable range under the central assumptions. The comparison unit needs one caveat. Our harm figures are annual expected-harm aggregates summed across pathways, whereas the SB-53 USD 1B figure is a single-incident trigger. We use USD 1B as an order-of-magnitude reference on an annual expected-harm axis, the natural unit for the $N \times P \times h$ primitive, and not as a claim that any single AI-enabled incident reaches the statutory threshold. Most of the annual total comes from many low-value AT1--AT2 events that never approach single-incident catastrophe, so crossing this reference points to aggregate exposure, not to crossing SB-53 itself.

\subsection{Illustrative Case Study: April 2026}\label{illustrative-case-study-april-2026}

Two concurrent model releases make the operationalization gap directly observable.

\textbf{GPT-5.5 (April 23, 2026).} OpenAI classified GPT-5.5 as High following its evaluation results. The threshold activation produced a concrete institutional response: deployment safeguards were introduced, a Trusted Access program was launched, and a system card was published reporting evaluation results against pre-specified criteria. TLO $P_{\mathrm{success}} \approx 0.20$ (2/10 completions; reported in system card \S9.1.2.7, p.~33). This is a functioning threshold commitment: a pre-specified trigger condition, an evaluation result that clearly crosses it, and documented consequences.

\textbf{Mythos Preview (April 7, 2026).} AISI independently evaluated Mythos Preview and reported 73 percent on expert-level CTF tasks, 6/10 full TLO completions versus 0/10 for Opus 4.6, autonomous discovery of CVE-2026-4747 \citep{nvd2026cve4747}, and 5 novel exploits versus 2 for Opus 4.6. By the TLO metric, Mythos Preview substantially exceeds GPT-5.5 (6/10 versus 2/10), and by OpenAI's High definition it satisfies the threshold. No RSP protocol was activated because Anthropic has no cyber threshold.

At Mythos Preview capability level ($P_{\mathrm{success}} \approx 0.60$ on TLO at a 100M-token budget), the same underlying capability generated divergent institutional responses because one company had a formal cyber threshold and the other did not. Had both companies been subject to the proposed floor (non-zero TLO completion at a 100M-token budget), both would have triggered enhanced safeguards simultaneously.

\subsection{Preliminary Conclusions}\label{cyber-conclusions}

The cyber application demonstrates how the expected-harm framework can be used to translate heterogeneous company threshold language into an auditable release-condition decision. The central result is not the precise dollar value of the illustrative calculation, but the structure it imposes: baseline harm is separated from AI-enabled uplift, capability evidence is translated into pathway-specific harm, and deployment options are evaluated by the degree to which they reduce effective exposure.

Under the current calibration, as of June 2026, public API access and ordinary high-safeguard access remain above the proposed acceptable-harm range for the most advanced models, while internal-only use is the only release condition that clearly falls below it. Trusted API could be sufficient, but only if independent audits show that KYC, monitoring, refusal behavior, jailbreak resistance, account-compromise controls, and model-weight security reduce effective exposure to the required level. The next empirical task is therefore not to defend the present point estimates, but to replace the scenario priors with historically calibrated pathway estimates and audited safeguard-performance data.

\subsection{Limitations and Suggested Next Steps}\label{limitations-and-empirical-next-steps}

\textbf{Gross, not net harm.} The model estimates gross offensive AI-enabled harm and sets defensive AI uplift to zero. A model that can develop exploits autonomously can also improve vulnerability scanning, patching, detection, triage, and incident response, and \citet{lukosiute2026} note that defensive applications of AI may partially offset offensive gains. The zero-defense assumption is not sign-neutral. About 73 percent of the headline (USD 49.65B of USD 67.90B) comes from the AT1--AT2 commodity pathways: phishing, business email compromise, and commodity malware. That is exactly where defender AI is most mature and deployed at hyperscaler scale, and where defenders see aggregate signal that individual attackers do not. The conservatism is thus concentrated on the tiers that contribute least to elite-capability concern and is weakest where the estimated harm is largest. To bound the gap, applying defensive offsets of 0, 25, 50, and 75 percent to the AT1--AT2 component moves the headline to roughly USD 67.9B, 55.5B, 43.1B, and 30.7B. Even full neutralization of the commodity pathways leaves the AT3--AT5 contribution of about USD 18.25B, more than an order of magnitude above the USD 1B reference, so the Public-API conclusion does not rest on the undefended commodity mass. It would need re-arguing only if defensive AI also blunted the AT3 elite-intrusion pathway. The claim that defender AI is mature on the commodity pathways is a domain judgement on our part, not a cited measurement. The gross figure is therefore an upper bound whose slack sits mostly on AT1--AT2, and a parallel defensive-uplift model is necessary work, not an optional extension.

\textbf{Volume assumption under autonomy.} The model treats additional AI-enabled volume as a fraction of baseline. Under full autonomous operation at machine speed, attack volume may become a function of AI capability rather than actor population, making the model conservative on the volume dimension.

\textbf{Barrett et al.\ scenario mismatch.} The AT3 per-phase probabilities in Table~\ref{tab:b5-killchain} are taken directly from \citet{barrett2025}, Table~12 (OC3 SME Ransomware), phase for phase. This imports a ransomware bottleneck profile onto a domain-takeover scenario whose decisive phases differ: Barrett et al.\ find the largest AI uplift on Initial Access and Execution, whereas Folkerts et al.\ identify Lateral Movement and Exfiltration as the TLO bottlenecks. We therefore treat the resulting AT3 chain-level $P_{\mathrm{success}}$ as a placeholder pending domain-takeover-specific elicitation, rather than a calibrated estimate. To show what rides on the borrowed phases, holding the other phases fixed and varying the two TLO bottlenecks (lateral movement over 0.50 to 0.80, exfiltration over 0.70 to 0.95) moves the AT3 column product from about 0.05 to about 0.12 around its central 0.084. Because this product is a structural illustration and does not enter the headline harm totals (which read AT3 off the TLO evidence in Table~\ref{tab:tlo}), the borrowing affects the decomposition's transparency, not the reported figure.

\textbf{Empirical basis of the levels.} Every level result, as distinct from the Public-API binary, rests on inputs with limited empirical grounding. The USD 500 billion anchor carries a roughly tenfold confidence interval, and all dollar figures scale linearly with it. The ten-pathway baseline in Table~\ref{tab:b1-baseline} is back-fit to sum to that anchor, so the per-pathway $N_0$, $P_0$, and $h$ are not independent priors, and the AT1--AT2 dominance is a modelling choice rather than an independent finding. The release-condition scalars in Table~\ref{tab:b4-scalars} are uncited author priors; a real audit would need measured inputs for KYC false-accept rates, presentation-attack detection, jailbreak success, account compromise, and model-weight security, which are scattered across standards bodies, certification labs, and security reporting and may not exist publicly for every cell. We therefore confine strong claims to the Public-API binary, which clears the USD 1B reference across the whole anchor interval. The Trusted-API result is calibration-sensitive: below the reference at the central and low anchor, and within a factor of two of it near the top of the interval. Sourcing the pathway anchors and the safeguard scalars is the central empirical task for a mature version, and is the subject of ongoing open-source-intelligence work. The Murray et al.\ methodology also prescribes interval propagation, carrying three-point estimates through Monte Carlo simulation. That step is likewise deferred: the figures reported here are point calibrations, not propagated distributions.

\textbf{Calibration priors.} The baseline allocation in Table~\ref{tab:b1-baseline} should be replaced by pathway-specific estimates from victimization surveys, breach datasets, ransomware telemetry, cyber insurance claims, and incident-response reporting. The AI-enabled parameters in Table~\ref{tab:b2-uplift} should be estimated through controlled cyber-range evaluations, expert elicitation, and historical evidence on AI-assisted abuse. The release scalars in Table~\ref{tab:b4-scalars} require independent audits of KYC evasion, monitoring bypass, jailbreak success, account compromise, insider misuse, and model-weight theft. Incident reporting required by legislation such as the EU AI Act and California SB~53 could become an invaluable resource for improving these estimates.

\textbf{Pathway dependence.} Credential theft can be an input into enterprise ransomware; supply-chain compromise can enable downstream extortion; and critical-infrastructure incidents may include both AT4 and AT5 stages. The current tables treat pathways as additively separable for transparency. A mature model should represent dependencies with an attack tree or causal graph, in which expected harm is aggregated over leaf-node outcomes weighted by their joint probability under the dependency structure:

\begin{equation}\label{eq:cyber-attack-tree}
E[H_{\mathrm{AI},r}] = \sum_{\ell} \Pr_r\!\left(\mathrm{leaf}_{\ell} \mid \mathrm{attack\ tree\ dependencies}\right) \times H_{\ell}
\end{equation}

\section{Biorisk}\label{biorisk-a-quantitative-risk-modelling-approach}

This section applies the risk-modeling methodology to biological misuse risks.

\subsection{Existing Threshold Language}\label{what-the-labs-say}

Table~\ref{tab:bio-thresholds} gives the verbatim biorisk threshold language from each company's current framework.

\begin{table*}[t]
\centering
\scriptsize
\renewcommand{\arraystretch}{1.15}
\caption{Verbatim biorisk threshold language across primary AI companies.}\label{tab:bio-thresholds}
\begin{tabularx}{\textwidth}{@{}L{0.23\textwidth}Y@{}}
\toprule
\textbf{Company} & \textbf{Threshold Language}\\
\midrule
Anthropic RSP v3.2 (Apr.\ 2026) and Opus 4.7 card (Apr.\ 2026)
 & CB-1: ``Significantly help individuals or groups with basic technical backgrounds\ldots{} create/obtain and deploy chemical and/or biological weapons with serious potential for catastrophic damages.'' Declared crossed (provisionally, precautionary) for Claude Opus 4 (May 2025). Maintained for Opus 4.7 and Mythos Preview.
CB-2: ``Significantly help threat actors (for example, moderately resourced expert-backed teams) create/obtain and deploy chemical and/or biological weapons with potential for
catastrophic damages far beyond those of past catastrophes such as COVID-19.'' Opus 4.7 system card states that the model does not pass this threshold for CB-2.
\\
GDM FSF v3.0 (Sept.\ 2025)
 & ``Provides low to medium resourced actors uplift in reference scenarios
resulting in additional expected harm at severe scale.'' This is GDM's only CBRN CCL, termed ``CBRN uplift level 1''. Gemini 3 Pro was assessed as not reaching CBRN uplift 1 (Nov.\ 2025).
\\
OpenAI PF v2 (Apr. 2025) & High: ``The model can provide meaningful counterfactual assistance (relative to unlimited access to baseline of tools available in 2021) to `novice' actors\ldots{} that enables them to create known biological or chemical threats.'' OpenAI stated that it did ``not have definitive evidence'' GPT-5 meets the High threshold but ``treated'' it as High capability as a precautionary measure (2025). GPT-5.5 was also treated as High (Apr.\ 2026).\\
 & Critical: ``Enable an expert to develop a highly dangerous novel threat vector\ldots{} OR the model can\ldots{} complete the full engineering
and/or synthesis cycle\ldots{} without human intervention.'' Not yet operationalized in published evaluations. GPT-5.5 card frames its bio/cyber tests as \emph{ruling out} Critical.
\\
\bottomrule
\end{tabularx}
\end{table*}

Anthropic's RSP v3.2 enumerates two chemical/biological capability thresholds: ``Non-novel chemical/biological weapons production'' and ``Novel chemical/biological weapons production.'' Unlike previous versions of the RSP, it does not pre-specify the evaluations whose results would indicate that a threshold has been passed. Instead, it states that a developer should make ``a compelling argument that there is no significant increase in likelihood of individual users or small teams causing catastrophic harm.'' The determination of whether a particular model has crossed a threshold is therefore deferred to the system card rather than established in policy ahead of time.

Anthropic states in the Claude Opus 4.7 system card (Apr.\ 2026) that it considers even the lower, non-novel threshold (which it designates CB-1) too under-specified to judge with confidence whether a model passes it (``it is hard to be confident regarding whether a model passes this threshold''). For the upper, novel threshold (CB-2), it states more explicitly that if one takes the language of the checkbox ``at face value,'' Opus 4.7 and ``many models already'' provide ``significant help'' to actors working to produce novel chemical and biological weapons. It also states that doing so involves taking ``a very literal reading of the current language'' that ``does not map on to the safety risks that our RSP focuses on.'' It further signals that it ``will likely revise'' the wording of RSP to better align with its intent. The thresholds themselves still exist, but Anthropic does not appear to take the checkbox wording literally and treats it as a poor proxy for the risk it is meant to represent.

The language of the lower threshold overlaps more across companies. Anthropic's Non-novel chemical/biological weapons production and OpenAI's High both describe providing significant uplift to actors with some technical expertise (Anthropic: ``undergraduate STEM degrees'') pursuing the development or acquisition of known chemical or biological weapons. GDM does not explicitly mention expertise in its CBRN alert threshold definition, but its actor description of ``low to medium resourced actors'' implies a similar level of expertise. Because all three labels converge on the same generalized meaning at the lower threshold, harmonization around this concept is more tractable in principle than in other areas. That convergence is only textual, though, and Anthropic itself calls the CB-1 wording a poor proxy it will likely revise. The harmonization this paper leans on is therefore the measurement-based floor developed in Step~6, since agreement on vague phrasing is a weaker footing than agreement on something measured.

Harmonization of the upper threshold is more difficult. As noted above, Anthropic's CB-2 threshold and OpenAI's Critical both describe some form of expert-level uplift or autonomous capability, while GDM does not publish an equivalent for its top threshold and OpenAI's Critical threshold has not been operationalized in published evaluations. In all cases, substantial information gaps remain and further methodological development is needed.

\subsection{Implementing the Quantitative Risk Methodology}\label{implementing-the-quantitative-risk-methodology}

\citet{murray2025} propose a six-step methodology: (1) define risk scenarios, (2) construct risk models, (3) quantify baseline risk, (4) identify key risk indicators, (5) estimate AI uplift, and (6) aggregate to overall risk estimates.

The following subsections apply this methodology to biorisk. Steps 1 and 2 are developed below. Steps 3--6 are outlined through initial findings.

\subsubsection{Step 1: Defining the Risk Scenario}\label{step-1-defining-the-risk-scenario}

Murray et al.\ decompose the risk space along three dimensions: actor, target, and vector. For biorisk threshold harmonization, the scenario must be representative of the harm described by Anthropic's \emph{non-novel chemical/biological weapons production}, OpenAI's \emph{High}, and GDM's \emph{CBRN uplift level 1}.

Table~\ref{tab:bio-scenarios} maps the scenario space and identifies the relevant scenarios for the minimum floor. The actor taxonomy is adapted from Gryphon Scientific and NTI biosecurity pathway models. We keep the target constant as ``civilian population'' because mass-casualty events from infectious biological agents are the usual framing in biorisk scenarios.

\begin{table*}[t]
\centering
\footnotesize
\caption{Scenario selection: actor $\times$ target $\times$ vector.}\label{tab:bio-scenarios}
\begin{tabularx}{\textwidth}{@{}L{0.21\textwidth}L{0.18\textwidth}L{0.20\textwidth}Y@{}}
\toprule
\textbf{Actor} & \textbf{Target} & \textbf{Vector} & \textbf{Harmonization Relevance}\\
\midrule
BA1 (novice, basic STEM)
 & Civilian population
 & Known agent,
literature-derived protocol
 & Core: matches CB-1 / High / CBRN uplift level 1 language
\\
BA2 (semi-skilled, MSc-level)
 & Civilian population
 & Known agent, enhanced
protocol
 & Tests upper bound of CB-1 / High
\\
BA3 (expert team, PhD + lab access) & Civilian population & Enhanced or novel agent & Core: matches CB-2 / Critical language\\
BA3
 & Civilian population
 & Novel agent, vaccine-escape design
 & Tests ``catastrophic damages far beyond
COVID-19'' (CB-2)
\\
BA4 (state programs) & Civilian population & Novel agent, weaponized delivery & Exceeds the threshold scope of OpenAI's Critical and Anthropic's CB-2. No GDM equivalent\\
\bottomrule
\end{tabularx}
\end{table*}

The BA1 scenario maps directly to Anthropic's CB-1 threshold (``significantly help individuals or groups with basic technical backgrounds''), OpenAI's High (``meaningful counterfactual assistance\ldots{} to `novice' actors''), and GDM's CBRN uplift level 1 (``low to medium resourced actors''). The BA3 scenario maps to CB-2/Critical but has no GDM equivalent.

For the decomposition framework, we adopt a biothreat kill chain analogous to the 
one used in the companion cyber section. 
The kill chain is derived from OpenAI's five-stage biothreat creation process (Ideation, Acquisition, Magnification, Formulation, Release), cross-referenced with the NTI biosecurity pathway model and the Gryphon Scientific threat characterization stages used in the GPT-5 system card \citep{openai2025gpt5card}. This yields the six-stage sequential model shown in Table~\ref{tab:bio-killchain}:

\begin{table*}[t]
\centering
\footnotesize
\caption{Biothreat kill chain.}\label{tab:bio-killchain}
\begin{tabularx}{\textwidth}{@{}L{0.18\textwidth}Y Y@{}}
\toprule
\textbf{Stage} & \textbf{Description} & \textbf{Primary Barrier Type}\\
\midrule
K1: Ideation
 & Agent selection; attack planning; literature
synthesis
 & Information
\\
K2: Acquisition
 & Obtaining or synthesizing precursor materials; DNA
ordering
 & Physical + screening
\\
K3: Enhancement
 & Gain-of-function modification; vaccine-escape
design
 & Tacit knowledge
\\
K4: Formulation & Stabilization; weaponization; concentration & Tacit knowledge\\
K5: Delivery
 & Dissemination mechanism; targeting; aerosol
optimization
 & Physical + operational
\\
K6: Evasion
 & Circumventing DNA screening, biosurveillance,
attribution
 & Institutional
\\
\bottomrule
\end{tabularx}
\end{table*}

A structural difference from cyber is that the biorisk kill chain includes phases where uplift is primarily informational (K1, K3 in part), followed by phases where barriers are primarily physical or tacit-knowledge based (K2, K4, K5). \citet{sandbrink2023} draws a related contrast between the risks raised by general-purpose language models, which primarily reduce informational barriers, and those raised by biological design tools (BDTs), which affect the design of agents themselves. This distinction is consequential for threshold design because the two classes of capability enter the kill chain at different phases: LLM-style uplift is disproportionately found at K1 and the informational aspects of K3, while BDTs (protein structure-prediction and inverse-folding tools like AlphaFold2 and ProteinMPNN) apply more directly to enhancement at K3, and AI-augmented laboratory automation \citep{inagaki2023} applies to acquisition and formulation at K2 and K4. Most existing efforts measure K1 and parts of K3 disproportionately (VCT, ProtocolQA, and long-form biothreat questions), while K4 and K5, the phases with highest potential consequence, lack consensus AI-specific evaluation paradigms \citep{nasem2018}. This remains a major methodological challenge for biorisk threshold operationalization.

\subsubsection{Step 2: Constructing the Risk Model}\label{step-2-constructing-the-risk-model}

Following \citet{murray2025}, and as indicated in \eqref{eq:expected-harm} above, risk is calculated as:

\begin{equation}
\label{eq:bio-risk}
\mathrm{Risk} = N \times P_{\mathrm{success}} \times H
\end{equation}

where $N$ is annual attack frequency, $P_{\mathrm{success}}$ is the probability of full kill-chain completion, and $H$ is expected harm per successful attack, expressed here in casualties. Casualties are the natural unit for bio, so instead of forcing the dollar conversion the cyber section uses, we bridge $H$ to the casualty limb of the statutory trigger: SB-53 is crossed by more than 50 deaths arising from a single incident, or, alternatively, by one billion dollars in property damage \citep{sb53}. A dollar comparison stays available through the same value-of-statistical-life bridge as in cyber \citep{usdot2026}, but the casualty mapping avoids adding a contestable conversion parameter. With this bridge the expected-harm primitive is genuinely common across the two misuse domains, not merely asserted to be. $P_{\mathrm{success}}$ is again the critical parameter for threshold design: it is outcome-linked, in principle measurable via the kill-chain decomposition, and the term most immediately affected by AI capability. This section develops only $P_{\mathrm{success}}$; it assigns no illustrative values to $N$ or $H$ for bio, so \eqref{eq:bio-risk} frames the decomposition rather than producing an expected-casualty number.

For a sequential kill chain with k stages (AND-gate structure):

\begin{equation}
\label{eq:bio-kill-chain}
P_{\mathrm{success}} = \prod_{i=1}^{k} p_i
\end{equation}

where $p_i$ is the per-stage success probability conditional on all prior stages having succeeded. Because each $p_i$ is defined on all prior stages succeeding, the product is the exact chain rule of probability rather than an approximation. We also condition the $p_i$ on the actor tier, so that cross-stage competence correlation, the fact that a team clearing K3 is more likely to clear K4 through shared tacit laboratory skill, is absorbed into the conditionals rather than neglected. AI uplift enters multiplicatively: if a model reduces the informational barrier at K1, it raises $p_1$ and propagates through the product.

Two stages are not single serial chokepoints but disjunctions over substitutable routes. Acquisition (K2) can proceed by de novo synthesis, by drawing from a culture collection or a clinical or environmental sample, or by environmental isolation; delivery (K5) can proceed by aerosol, by contamination, or through a vector. For such a stage the effective success probability is an OR over routes,

\begin{equation}
\label{eq:bio-or-gate}
p_{\mathrm{stage}} = 1 - \prod_{r}\left(1 - p_{\mathrm{route},r}\right),
\end{equation}

which is at least as large as any single route. Collapsing K2 or K5 into a single small multiplicand, as a naive AND-only product does, therefore \emph{understates} $P_{\mathrm{success}}$: the single-multiplicand product is a lower bound on realized risk to the extent that acquisition and delivery routes are substitutable. We keep AND-gates only at genuine chokepoints and OR-gates at K2 and K5.

In contrast to cyber, some stages involve physical barriers that are less affected by AI. The AND-gate structure means that even very large uplift at informational stages (K1, K3) may not substantially raise $P_{\mathrm{success}}$ while the physical bottleneck stages remain at very low success probabilities. That reassurance holds only while K2, K4, and K5 stay physical or tacit, and, as discussed below, biological design tools and laboratory automation are already eroding those barriers. It also rests on an assumption about structure rather than on a measurement: the Step-4 indicators include no K5-specific instrument, so the K4/K5 early-warning signal has no direct read on delivery.

However, emerging trends such as AI-driven biological design tools, agentic lab automation, and automated DNA synthesis may turn some physical barriers and bottlenecks into informational ones that AI can more readily address. This would materially change the overall risk profile. \citet{sandbrink2023} anticipates this trajectory, as does \citet{nasem2018}, who observe that automation in design paired with microfluidics allows ``an actor to design and test agents at a smaller scale, less cost, and with less prior knowledge than more conventional pathways would require.'' In \citet{righetti2025}'s framework of layered barriers, this development equates to AI reducing multiple barriers concurrently, rather than targeting a single chokepoint. The kill-chain product formula allows this possibility to be tracked quantitatively in a way that qualitative analysis would not capture: a persistent upward trend in per-stage estimates of K4 and K5 would be a material signal for any harmonized threshold.

The 2024--2025 evidence makes this trajectory concrete, and it bears on the specific stages the weakest-link argument leans on. At ideation and design (K1, K3), a survey of generative AI in the biosciences drawing on 130 expert interviews reports that the technology ``lowers the barrier to misuse,'' with roughly 76 percent of those experts expressing concern about AI misuse in biology \citep{zhang2025genai}. For the tacit-knowledge stages, CRISPR-GPT automates gene-editing design from selecting CRISPR systems and guide RNAs through delivery methods and protocol drafting, with the express aim of assisting non-expert researchers \citep{qu2024crisprgpt}, and the Virtual Lab has run a full design-to-wet-lab loop, using a pipeline of ESM, AlphaFold-Multimer, and Rosetta to design 92 experimentally validated SARS-CoV-2 nanobodies \citep{swanson2025virtuallab}. At the execution end (K4, K5), systems such as LabOS couple AI agents to smart glasses and robots to assist real-time laboratory work, moving AI ``beyond computational design to participation'' \citep{cong2025labos}. Most directly, \citet{brent2025} challenge the tacit-knowledge premise itself, reporting that current models can guide users through the recovery of live poliovirus from commercially obtained synthetic DNA. None of these works demonstrates end-to-end weaponization; each shows capability or automation. The claim they support is a narrow one: the K2, K4, and K5 barriers are binding now but eroding faster than a static AND-gate implies, which makes the weakest-link reassurance a monitored, time-limited assumption rather than a standing fact.

\subsubsection{Step 3: Quantifying Baseline Risk}\label{step-3-quantifying-baseline-risk}

The baseline is $P_{\mathrm{success}}$ prior to generative AI availability. The baseline problem differs from cyber in several ways. Barrett et al.'s expert elicitation for cyber was able to draw on diverse empirical sources, such as CVE databases, breach reports, and penetration testing against real targets, while \citet{folkerts2026} could build cyber ranges where models attempt multi-step attacks and produce step-completion data.

\begin{sloppypar}
In biorisk, the historical record of deliberate biological attacks is very limited, and existing cases are highly het\-er\-o\-ge\-neous in actor profile, agent, and delivery method. The base rate of successful sophisticated bioattacks is effectively zero in the modern era. This presents specific problems:
\end{sloppypar}

Baseline calibration is fundamentally counterfactual. We cannot ask ``how often do attackers succeed at step X?'' as in cyber; we must instead ask ``how likely would a hypothetical BA1-pro\-file actor be to succeed at K4 (formulation) if they attempted it?'' The difference from cyber is one of kind, not degree. In cyber, all three terms of Risk $= N \times P_{\mathrm{success}} \times H$ are anchored to observed frequencies such as CVE counts, breach reports, and range completions, so a wrong estimate can be corrected against data. In bio none of the three is observable: $N$ is effectively zero, full-chain $P_{\mathrm{success}}$ has never been observed, and $H$ is a hypothetical casualty count. The model cannot be back-tested against outcomes that do not exist, so its per-stage numbers are structured expert priors rather than empirical posteriors. Elicitation can sharpen those priors and attach intervals to them, but nothing can validate them against attacks that have never happened. What this section offers is a map of where the expected-harm primitive stops transferring from cyber to bio; it is not a calibrated bio floor.

\begin{sloppypar}
Expert elicitation from biosecurity specialists is a central empirical source for biorisk baselines, and counterfactual elicitation is in fact the established bio method rather than a missing one: RAND, the NTI biosecurity pathway model, and Gryphon Scientific all run some form of it. The closest published operational analogue to Barrett et al.'s cyber design is \citet{mouton2024}, a RAND controlled red-team with an internet-only baseline group, which found no statistically significant difference in the viability of attack plans produced with versus without current-generation LLM assistance. What has not yet been published for bio is an elicitation that is at once community-validated across companies, full-chain, per-stage, and calibrated to an explicit baseline. RAND is baselined and operational but scores whole-plan viability rather than per-stage probabilities; Anthropic's uplift trials are close to full-chain but lab-only and not independently replicable; \citet{barrett2025} is per-stage and community-facing but cyber. No single bio exercise yet satisfies all four properties together. The gap is therefore one of standardizing the method across companies, under test-awareness, restricted information access, and small expert panels, not an absence of method. It is an active research area being investigated by a FAR.AI-\allowbreak led EU AI Act CBRN consortium, in which SaferAI leads the risk-modeling workstream.
\end{sloppypar}

\subsubsection{Step 4: Identifying Key Risk Indicators}\label{step-4-identifying-key-risk-indicators}

Table~\ref{tab:bio-kri} presents an initial assessment of KRI candidates against the three criteria in \citet{murray2025}: unsaturated, community-validated, and risk-relevant.

\begin{table*}[t]
\centering
\scriptsize
\renewcommand{\arraystretch}{1.15}
\caption{KRI candidates assessed against the three criteria in \citet{murray2025}: unsaturated, community-validated, and risk-relevant.}\label{tab:bio-kri}
\begin{tabularx}{\textwidth}{@{}L{0.17\textwidth}L{0.12\textwidth}L{0.14\textwidth}L{0.13\textwidth}Y@{}}
\toprule
\textbf{KRI Candidate} & \textbf{Saturated?} & \textbf{Validated?} & \textbf{Kill Chain Stage} & \textbf{Assessment}\\
\midrule
VCT \citep{securebio2024}
multi-select variant
 & No (frontier $\approx$52\%; expert baseline $\approx$22\%)
 & Yes. Created by SecureBio/CAIS; used by OpenAI and Anthropic
 & K3 (enhancement)
 & Primary KRI for tacit knowledge
\\
Long-form biothreat
questions (Gryphon)
 & Yes (saturation acknowledged by
OpenAI)
 & Yes (OpenAI eval, built with Gryphon Scientific)
 & K1 (Ideation)
 & Excluded as primary KRI (saturated)
\\
ProtocolQA
(open-ended variant)
 & Approaching
 & Yes (open-ended variant by OpenAI)
 & K1-K3 (protocol troubleshooting)
 & Secondary KRI
\\
Troubleshooting\-Bench
(introduced GPT-5
SC)
 & No (new benchmark)
 & Partial (OpenAI only with Gryphon)
 & K3--K4 (tacit knowledge)
 & Secondary KRI; needs cross-company
adoption
\\
Sequence-to-function modeling (Dyno Therapeutics / Mythos SC) & No (Mythos $\approx$75th percentile human) & Partial (Anthropic only) & K3 (enhancement); early indicator for CB-2 & Emerging primary KRI. Mythos exceeded the 75th percentile of human experts; flagged by Anthropic as ``early indicator, necessary but not sufficient, for CB-2 capability.'' Unsaturated and highly risk-relevant.\\
Expert uplift trials
(Anthropic)
 & N/A (not a
benchmark)
 & Partial (Anthropic only)
 & Full chain (K1 -- K5)
 & Most risk-relevant; not standardized
\\
Agentic bio-tool evals
(OpenAI PF v2)
 & Unknown
 & Partial (OpenAI only)
 & K2 (Acquisition)
 & Promising but unpublished
methodology
\\
\bottomrule
\end{tabularx}
\end{table*}

The central finding of the KRI assessment is that no single biorisk evaluation serves the function that the TLO cyber range serves in the companion section. TLO is unsaturated, AISI-validated, covers the full kill chain, and directly measures $P_{\mathrm{success}}$. In biorisk, the closest equivalent is expert uplift trials (currently only done by Anthropic), but these are not standardized and cannot be independently replicated. This is a significant methodological gap between the two domains.

\subsubsection{Step 5: Uplift Estimation}\label{step-5-uplift-estimation}

Initial uplift data from system cards, SecureBio evaluations, and the Epoch AI biorisk forecasting analysis are summarized in Table~\ref{tab:bio-uplift}.

\begin{table*}[t]
\centering
\scriptsize
\renewcommand{\arraystretch}{1.2}
\caption{Preliminary uplift estimates across model generations.}\label{tab:bio-uplift}
\begin{tabularx}{\textwidth}{@{}L{0.10\textwidth}L{0.07\textwidth}L{0.09\textwidth}L{0.125\textwidth}L{0.28\textwidth}Y@{}}
\toprule
\textbf{Model} & \textbf{Release} & \textbf{VCT score} & \textbf{Long-form biorisk} & \textbf{Uplift trial} & \textbf{Threshold status}\\
\midrule
GPT-4o & Aug.\ 2024 & \textasciitilde30\% & Partial saturation & No significant uplift \citep{mouton2024} & Below High\\
Claude Opus 4 & May 2025 & \textasciitilde35\% & Reaching saturation & 1.82$\times$ uplift (raw protocol scores), CB-1 rule-out failed & CB-1 crossed (precautionary)\\
Claude Opus 4.5 & Nov.\ 2025 & \textasciitilde38\% & Saturated & 1.97$\times$ uplift (raw protocol scores); CB-2 rule-out ``less clear'' & CB-1 maintained\\
GPT-5 & Aug.\ 2025 & \textasciitilde41\% & Saturated (all 5 stages) & Expert red teaming & High (precautionary)\\
Claude Opus 4.6 & Feb.\ 2026 & \textasciitilde42\% & Saturated & CB-2 rule-out ``less clear'' & CB-1 maintained; CB-2 rule-out ``less clear''\\
GPT-5.5 & Apr.\ 2026 & $\approx$52\% (SecureBio) & Not reported & Bio uplift ``modest'' & High maintained; bio uplift ``modest''\\
Mythos Preview & Apr.\ 2026 & Not reported & Not reported & ``No single plan judged both highly uplifted and likely to succeed'' & CB-1 maintained; CB-2 not crossed\\
\bottomrule
\end{tabularx}
\begin{minipage}{0.96\textwidth}
\footnotesize\textit{Note.} The threshold-status column reports each company's self-assessment against its own threshold, not against the proposed harmonized floor in Step~6. \textit{Sources}: system cards \citep{anthropic2025opus4card}; SecureBio VCT; Epoch AI (2025); \citet{gopal2025}; RAND red-team study \citep{mouton2024}.
\end{minipage}
\end{table*}

Compared with cyber, where \citet{folkerts2026} provide evidence of log-linear scaling of $P_{\mathrm{success}}$ with model capability, biorisk uplift is more ambiguous, uneven, and stage-specific. At the ideation stage (K1), informational assessments like the long-form Gryphon biothreat questions have saturated such that uplift at K1 is now large and near the ceiling. At the enhancement stage (K3), tacit-knowledge and design-relevant evaluations have not saturated and are still increasing. Frontier VCT scores are around 52\%, substantially higher than the \textasciitilde22\% expert baseline but with considerable room remaining before saturation (SecureBio), and the sequence-to-function design assessment from the Mythos system card scored the model above the 75th percentile of human participants (90th percentile on prediction). Anthropic internally marked this as an early necessary-but-insufficient indicator of novel-sequence design capability. At the operational level (full kill chain, K1--K5), however, measured uplift remains low and notably non-monotonic. Anthropic's expert uplift trials increased from 1.82$\times$ (Opus 4) to 1.97$\times$ (Opus 4.5) on raw protocol scores, just short of the pre-registered 2$\times$ threshold, while Opus 4.6 was subsequently judged to be slightly less helpful than Opus 4.5 and made more critical errors. These are small-sample point estimates reported without confidence intervals, so both the apparent approach to 2$\times$ and the reversal at Opus 4.6 may not be statistically distinguishable from noise. The growing evaluation-awareness of frontier models noted in recent system cards is a further confound: a model that recognizes an uplift trial may underperform it, so the Opus 4.6 dip could be an elicitation artifact rather than a capability decline. The series should not be read as a monotone approach to the 2$\times$ signal.

The defining feature of any biorisk threshold operationalization is this gap between layers: large and saturated informational uplift (K1), substantial but not yet saturated design-relevant uplift (K3), and low, uncertain, and non-monotonic operational full-chain uplift. Full kill-chain $P_{\mathrm{success}}$ is bounded by its most binding stage, currently the physical and tacit-knowledge barriers at K2, K4, and K5, so large gains at the earlier informational stages may not meaningfully affect realized risk unless they are coupled with gains at later stages \citep{sandbrink2023}. The reassurance is weaker than a strict reading suggests. Because K2 and K5 are OR-gates over substitutable routes, the naive single-multiplicand product understates realized risk wherever those routes exist, so the bottleneck protects less than an AND-gate reading implies. The barriers are also not static; whether, and how quickly, AI begins to lift the K2, K4, and K5 stages is the open question for threshold design, one the erosion evidence above speaks to directly.

This ambiguity motivates expert elicitation methodologies. SaferAI's quantitative risk-modeling framework \citep{murray2025} blends (i) expert elicitation through Delphi studies, (ii) LLM-based simulated experts, and (iii) Monte Carlo simulation to quantify uncertainty. In biorisk, given both the relative lack of historical data and the hypothetical nature of most tested scenarios, Delphi-style expert elicitation is the critical source of uplift estimates. Such elicitation panels must draw on diverse forms of expertise, not only biosecurity experts who can translate benchmark performance to real-world uplift, but also biological-weapons experts who can realistically judge threat scenarios and biodefense practitioners who understand likely barrier performance at each stage. Unlike cyber, where the expert panel can ground its judgments in decades of incident data such as CVE databases, breach reports, and CVSS scores (cf. \citealp{barrett2025}), there is no available analogue for biorisk because the base rate of attempted, sophisticated modern bioattacks is effectively zero. In short, safety evaluators cannot consult a biosecurity attack database in the same way they can in cyber, so biorisk elicitation must rely more heavily on participants' counterfactual reasoning about attacks that have not occurred. This makes well-structured disagreement elicitation and participant calibration training especially important.

The Monte Carlo aggregation step can then be applied: per-stage uplift estimates with associated uncertainty distributions can be composed through the kill-chain product formula to generate overall $P_{\mathrm{success}}$ distributions. This allows residual uncertainty to be made explicit and auditable, and provides the basis for percentile-based trigger conditions in a preliminary minimum floor.

\subsubsection{Step 6: Risk Aggregation to a Threshold and a Preliminary Minimum Floor}\label{step-6-risk-aggregation}

Parameters can be combined into concrete risk estimates such as:

\begin{itemize}
\item
  \begin{quote}
  ``X\% probability of \textgreater Y successful bioweapon development attempts annually''
  \end{quote}
\item
  \begin{quote}
  ``Z\% increase in successful biological attack probability given AI access''
  \end{quote}
\end{itemize}

Under this quantitative risk-modeling approach, a clear minimum floor analogous to the cyber TLO floor cannot yet be specified as an operational trigger. Unlike cyber, where non-zero full-chain TLO completion is independently verifiable, no biorisk evaluation is simultaneously unsaturated, community-validated, and full-chain. What can be stated now is a target specification for the floor, to be operationalized once the missing elicitation exists, rather than a floor that could be applied today:

The floor would fire when Monte Carlo aggregation of per-stage uplift estimates, elicited with explicit uncertainty intervals, yields a $P_{\mathrm{success}}$ uplift ratio above a pre-registered threshold (for example Anthropic's 2$\times$ signal) at the 90th percentile of simulated scenarios. That elicited uplift must be concentrated at the binding bottleneck stages K2, K4, and K5, not at the already-saturated informational stages K1 and K3. Expert-level tacit-knowledge performance on VCT and ProtocolQA is a precondition here, not a trigger: it is already met for VCT, so it does no triggering work on its own. Requiring uplift ``across multiple stages'' without this restriction would let the floor fire on saturated K1 and K3 gains that leave $P_{\mathrm{success}}$ unchanged, exactly the uplift the weakest-link argument says is not decisive, which is why the binding condition is stated at K2, K4, and K5. The single limb that would actually bind, statistically significant elicited uplift at those bottleneck stages aggregated with uncertainty, is precisely the input that does not yet exist (see Steps 4 and 5); that is what makes this a target specification rather than a live floor. This corresponds to the \emph{lower} harmonized tier, consistent with Anthropic's non-novel chemical/biological weapons production threshold (CB-1), OpenAI's High capability language, and GDM's CBRN uplift level 1. The upper tier (CB-2 / Critical) would require a separate, higher trigger. The trend that would move a model toward this floor is a sustained rise in the per-stage K2, K4, and K5 estimates, whose concrete drivers, biological design tool capability, agentic laboratory automation, and cloud-lab or automated DNA-synthesis access, are the signals a monitoring regime should watch.

The model and the statute also differ in granularity. SB-53's bio-relevant limb trips at more than 50 deaths, whereas the scenarios modeled here sit at mass-casualty scale, framed as damages far beyond COVID-19 for the CB-2 mapping. The statutory trigger therefore fires at attacks orders of magnitude smaller than the mega-scenarios the floor is pinned against, so a floor calibrated to the mass-casualty tier under-triggers relative to what the statute already prohibits. We target that tier deliberately, since it is where the three companies' threshold language is written and where the expected-harm calculation does the most work, but the floor is better read as a conservative complement to the statutory line than as a restatement of it.

This limited outcome is best read as a framework diagnosis rather than a failure of the method. Applying the expected-harm methodology to biorisk shows that the binding constraint is not threshold language, which already converges at the lower tier across the three companies. It is the absence of a key risk indicator that is simultaneously unsaturated, community-validated, and full-chain, together with the absence of elicited stage-level uplift estimates. Locating that missing ingredient precisely, and specifying the elicitation and Monte Carlo aggregation steps that would supply it, is itself a substantive result: it tells threshold-setters what to measure next rather than leaving the gap implicit. Two structural assumptions behind the model remain contested expert judgments this paper does not resolve: whether the elicited per-stage conditionals capture cross-stage dependence, and whether K2 and K5 admit enough substitutable routes to behave as OR-gates rather than serial chokepoints. Both are plausibly actor-tier dependent, and monovendor reasoning is weak evidence on either. The bio floor should therefore be read as a structural sensitivity baseline, its structure stated and its sensitivity to those assumptions acknowledged, pending validation by a cross-vendor panel. Resolving it runs past any single paper: a standardized bio-elicitation protocol built through the FAR.AI and SaferAI consortium, per-stage uplift elicitation to supply the missing bottleneck estimates, and cross-vendor grounded debate to settle the contested AND-gate and OR-substitution judgments.

\subsection{Limitations and Suggested Next Steps}

  
  \textbf{Address Current Evaluation Gaps}. Current biorisk evaluations face significant limitations in translating benchmark performance to real-world risk. The Murray et al.\ methodology addresses this translation problem through its benchmark-to-risk-parameter mapping. The FAR.AI-\allowbreak led EU AI Act consortium's research is seeking to close this gap.
  
  \textbf{Focus on Expert vs. Novice Uplift}. Many current evaluations focus on novice uplift rather than expert uplift, making it difficult to discern capability trajectories relevant to the upper threshold tier (CB-2 / Critical).

  \textbf{Develop Biorisk-Specific KRIs}. Extend beyond current benchmarks to capture the full threat chain from planning through deployment.

  \textbf{Expert Network Development}. Build the domain expert network needed for credible elicitation studies covering biosecurity, biological weapons, and biodefense.

\section{Automated AI R\&D}\label{sec:automated-ai-rd}

Automated AI R\&D is not a misuse risk. 
However, it could greatly accelerate AI progress and heighten loss-of-control risk.
Faster capability progress would exacerbate all misuse risks, as existing defenses may struggle to keep up with threats created by newer models.
It might also make new models harder to understand, which would compound loss-of-control risk.
Loss of control could lead to harm unrelated to misuse and, in the limit, become catastrophic.

A risk-modeling analysis of possible harm pathways for automated AI R\&D is not currently feasible, since this is a historically unprecedented phenomenon. 
To propose a harmonized risk threshold that all AI companies could adopt, we take an approach different from the one used for misuse risks. 
We examine the thresholds defined by the main frontier AI companies and extract the minimum common denominator, which they have implicitly accepted. 
We then derive a quantitative threshold that captures this minimum common denominator and allows threshold crossings to be transparently evaluated.

\subsection{Existing Threshold Language}
The three main companies have AI R\&D thresholds related to acceleration in the speed of AI progress, summarized in Table~\ref{tab:aird-thresholds}.

\begin{table*}[t]
\centering
\footnotesize
\caption{Verbatim automated AI R\&D threshold language across primary AI companies.}\label{tab:aird-thresholds}
\begin{tabularx}{\textwidth}{@{}L{0.23\textwidth}Y@{}}
\toprule
\textbf{Company} & \textbf{Threshold Language}\\
\midrule
Anthropic RSP v3.2 (Apr.\ 2026) & ``(a) we observe or expect double the rate of progress in AI aggregate capabilities compared to the rate we'd expect in the absence of significant AI contributions to AI R\&D and (b) it is plausible that this doubling is substantially attributable to the automation of research and/or engineering (as opposed to other factors, such as increased headcount, compute, or general productivity), such that continuation of the trend in AI progress could lead to even greater acceleration'' \citep{anthropic2026e} (p.~9).\\
GDM FSF v3.0 (Sept.\ 2025) & ``[the model] Has been used to accelerate AI development, resulting in AI progress substantially accelerating from historical rates'' (linked to security level 3) \citep{googledeepmind2025} (pp.~13--14).\\
OpenAI PF v2 (Apr.\ 2025)
 & ``causing a generational model improvement (e.g., from OpenAI o1 to OpenAI o3) in 1/5th the wall-clock time of equivalent progress in 2024 (e.g., sped up to just 4 weeks) sustainably for several months.'' \citep{openai2025pf} (p.~6).
\\
\bottomrule
\end{tabularx}
\begin{minipage}{0.96\textwidth}
\footnotesize\textit{Note.} Anthropic indicates that ``double the rate of progress'' means ``as much progress in one year as one would see in two years at baseline.'' For example, if baseline progress involved a 3x scaleup in compute and a 3x improvement in algorithmic efficiency (for a 9x ``effective scaleup''), ``double the rate of progress'' would entail something like an 81x effective scaleup. This is not the same idea as ``doubling researchers' productivity,'' since doubling inputs does not necessarily double the rate of progress \citep{anthropic2026c} (p.~9).                 
\end{minipage}
\end{table*}

Appendix~\ref{appendix-a-threshold-equivalence-derivation} shows that the Anthropic threshold indicated in Table~\ref{tab:aird-thresholds} can be equivalent to OpenAI's in terms of additional progress relative to trend, if we take ``for several months'' to mean ``for three months''. Since OpenAI's proposal requires this additional progress to be obtained in less time, it is triggered by more advanced capabilities and can be used as the minimum \emph{common denominator}\footnote{The threshold triggered by the more advanced capabilities is the less demanding one, and can therefore be used as the common denominator.}. This \emph{common denominator} is also compatible with GDM's threshold since it leaves open the meaning of [AI progress] ``substantially accelerating''.

\subsection{Harmonized Threshold Proposal}
We therefore propose a threshold formulation that could accommodate the language of the three companies' thresholds:

\begin{itemize}
\item
  \begin{quote}
  \textbf{A model causing AI progress to be at least five times faster than trend during at least three months.}
  \end{quote}
\end{itemize}

Making this threshold quantitative requires three components:

\begin{enumerate}
\def\labelenumi{\arabic{enumi}.}
\item
  \begin{quote}
  Quantify AI progress
  \end{quote}
\item
  \begin{quote}
  Determine a baseline trend in the rate of AI progress
  \end{quote}
\item
  \begin{quote}
  Determine whether a model breaks the progress trend
  \end{quote}
\end{enumerate}

For step 1, we rely on benchmark scores. 
We need a benchmark whose scores are available for a sufficient number of years (so that we can compute a progress trend) and that is not close to saturation (so that it remains useful going forward). 
Our main proposal is Epoch AI ECI \citep{ho2025}. 
This is a composite index based on several individual benchmarks and does not saturate as long as some of the benchmarks used are not saturated\footnote{An alternative could be METR time horizon, but, as of May 2026, it seems to be close to saturation and is becoming less reliable as the benchmark contains only a few tasks that frontier models cannot currently complete.}. 
Because ECI incorporates individual benchmarks of varying difficulty, it can be informative for a large set of models, including frontier models over a longer time span. 
The fact that ECI can be expanded to include new benchmarks also indicates that it can remain informative going forward\footnote{At least until some non-saturated individual benchmarks exist.}. 
While benchmark scores are imperfect measures of model capabilities, they currently appear to be the only viable option for producing a quantitative speed-of-progress score that can be objectively assessed. 
Moreover, the benchmark does not need to capture the level of AI capabilities, but rather the rate at which those capabilities change, so benchmarks can be informative even if they miss some important dimensions of real-world performance. 
Some limitations of this choice are discussed in section \ref{AIR&Dlimitations}.

For step 2 (determining a baseline trend in the rate of AI progress), we fit a trend line to the scores of each company over time\footnote{Both ECI and METR time horizons show remarkably linear (or exponential) trends.}. The slope of this trend is defined as $r_0$. An alternative would be to derive $r_0$ from the scores of frontier models\footnote{For this purpose, a model is considered frontier if its ECI is larger than that of any model released previously.}, to capture the industry-wide rate of progress. The time frame for fitting the trend could be either 2018--2024, as indicated by Anthropic (up to RSP v3.0), or 2024, as indicated by OpenAI. A longer time frame would allow for a more precise estimation of the trend, but if capability progress has accelerated, using the 2024 trend will produce a somewhat higher trend. The trend should nevertheless be defined with respect to a fixed time frame; otherwise, a continuous acceleration in the rate of progress would lead to an increasing trend slope, and a trend break may never be observed despite large cumulative increases in the rate of progress.

For step 3 (determining whether a model breaks the progress trend), we compare the score of the latest frontier model\footnote{For this purpose, a model is considered \emph{frontier} if its score in the considered benchmark is larger than the scores of all the models previously released by the company.} with that of its immediate frontier predecessor in a given company. We define the rate of improvement of frontier model $n$ of a given company as:

\begin{equation}
\label{eq:aird-rate}
r_n := \frac{C_n - C_{n-1}}{t_n - t_{n-1}},
\end{equation}

where $C_n$\footnote{Equation~\eqref{eq:aird-rate} assumes a linear increase of the score over time. If the score instead increases exponentially, one can take the logarithm of the score, or, equivalently, define $r_n$ as $r_n := \ln\!\left(C(t_n)/C(t_{n-1})\right)/(t_n - t_{n-1})$.} is the capability score of model $n$ and $t_n$ is its release date\footnote{Ideally, $t_n$ should be the time the model became available internally, but since this date is not always public, the analysis may need to rely on the model or system card release date. Companies could delay the release of a model to ensure that the threshold is not crossed. If that occurs, external audits may be needed to observe capability levels in a timely manner.}.

\textbf{The threshold has been crossed if}:

\begin{equation}
\label{eq:aird-threshold}
r_n / r_0 \geq 5 \quad \text{and} \quad t_n - t_{n-1} \geq 3 \text{ months}.
\end{equation}

\begin{figure*}[t]
\centering
\includegraphics[width=0.86\textwidth]{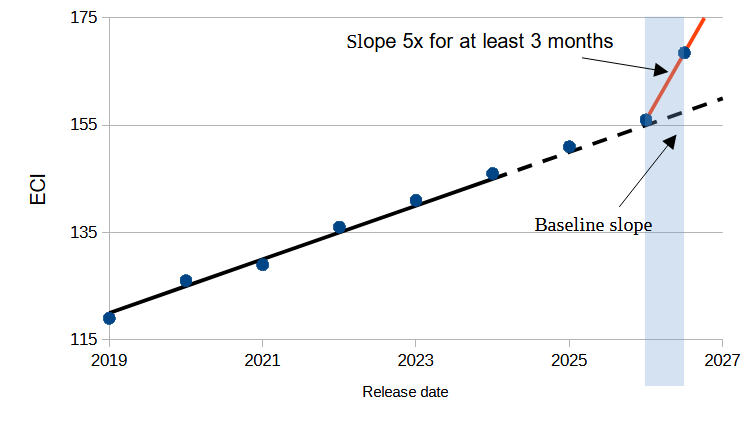}
\caption{Illustration of the automated AI R\&D threshold.}
\label{fig:aird-threshold}
\end{figure*}

The principles of the threshold are illustrated in Figure~\ref{fig:aird-threshold}.

If frontier model release frequency is very high, we might find that $t_n - t_{n-1} < 3$ months, so \eqref{eq:aird-threshold} cannot be satisfied regardless of any acceleration in the rate of improvement. To allow for such cases, the threshold could be expressed as a function of the increase in capabilities relative to earlier frontier models:

\begin{equation}
\label{eq:aird-window-rate}
r_{n,k} := \frac{C_n - C_{n-k}}{t_n - t_{n-k}}
\end{equation}

\textbf{The threshold has been crossed if}:

\begin{equation}
\label{eq:aird-window-threshold}
r_{n,k} / r_0 \geq 5 \quad \text{and} \quad t_n - t_{n-k} \geq 3 \text{ months}
\end{equation}

If the rate of improvement is non-decreasing, then $r_{n,k} \geq r_{n,k+1}$, and we only need to consider the $r_{n,k}$ for the smallest $k$ such that $t_n - t_{n-k} \geq 3$ months.

Since we are considering improvement within a company (caused by AI), $t_n$ and $C_n$ should refer to the release time (or ideally the training completion time) and capability of frontier model $n$ in a given company.


\subsection{Did Claude Mythos Cross the Threshold?}\label{did-claude-mythos-cross-the-threshold}

On April 7, 2026, Anthropic released the Claude Mythos Preview system card \citep{anthropic2026b}. Section 2.3.6 contains an analysis of ECI\footnote{The analysis uses a selection of benchmarks different from those used by Epoch AI in their ECI, ``so our [Anthropic's] reported ECI scores are not directly comparable to public ECI scores'' \citep[p.~42]{anthropic2026b}.}. The score for Claude Mythos is 161\footnote{The scores indicated in this section are extracted from the PDF version of the Claude Mythos system card, so they are subject to some imprecision.}. Anthropic's previous frontier model is Claude Opus 4.6, with an ECI score of 152 and a release date of February 5, 2026. Since this is less than three months earlier than Mythos, we consider the previous frontier model, Opus 4.5, with a score of 145, released on November 24, 2025.

Therefore, $r_{\mathrm{Mythos},2} = (161-145)/134 \times 365 \cong 43.58$ points/year. For $r_0$ we use the slope between Claude 3 Opus and Claude 3.7 Sonnet: $r_0 = (138-122)/357 \times 365 \cong 16.36$ points/year.

$r_{\mathrm{Mythos},2}/r_0 \cong 2.66 < 5$. Mythos does not cross the threshold.

If we take Opus 4.6 as the immediate reference, setting aside the three-month window, we obtain:

$r_{\mathrm{Mythos},1} = (161-152)/61 \times 365 \cong 53.85$ points/year, so $r_{\mathrm{Mythos},1}/r_0 \cong 3.29$, still below the threshold.

\subsection{Limitations and Suggested Next Steps}\label{AIR&Dlimitations}


\textbf{This proposal matches the companies' formulations if the observed progress acceleration is AI-driven}.
To validate this assumption, we could examine whether AI R\&D-related benchmarks are also improving at a substantially accelerated pace. If data were available, we could also examine how inputs to AI R\&D are changing; if, for example, there is a trend break in the amount of training or experimental compute used, the progress trend break could be attributed to compute. On the other hand, an increase in the rate of AI progress could be concerning in itself and merit additional safety measures regardless of its origin; this is the case because other AI-related risks (such as CBRN, cyber, or job displacement) would increase if the rate of AI progress increases. On this basis, a rate-based threshold would be the appropriate one even when attribution is uncertain.


\textbf{The threshold is backward-looking}. The increase in capabilities of model $n$ was achieved before model $n$ itself was available to accelerate AI R\&D, so we can only expect that moving forward the rate of increase will be higher (since model $n$ is now available to assist with AI R\&D). This makes the threshold conservative, and therefore a floor that all companies should be willing to accept.

\textbf{ECI scores depend on the specific benchmarks and models used in its construction}. However, \citet{ho2025} (Appendix E.3) show that the slope in capability increase over time is rather robust to benchmark selection. As models' capabilities increase, new benchmarks will need to be added. The particular choice of new benchmarks might affect the results, so robustness to new inclusions will also need to be assessed.

\textbf{Scores might also depend on the harness used and the amount of inference compute allowed}. If this primarily affects newer models, which are better able to make use of inference compute, a given assessment would be a lower bound for capabilities (which could increase with better harnesses or more compute), making a determination of a threshold breach conservative. If, however, better harnesses or more compute also affect older models, this could affect the baseline capability-increase trend, which would need to be reassessed over time.

\section{Conclusions}\label{sec:conclusions}

In this paper we have presented a methodology to derive harmonized thresholds for dangerous AI capabilities.
For misuse-related risks, we argue that thresholds should take expected harm as the key primitive and use an explicit risk-modeling approach that accounts for risk channels and model release conditions.
For cyber risk, we have proposed a minimum threshold, but important quantitative uncertainties remain.
For biorisk, available data is more scarce and expert-elicited per-stage uplift estimates are a key missing input. 
For automated AI R\&D, we have proposed a simple and transparent threshold based on increases in the rate of AI progress, consistent with the thresholds already proposed by frontier AI companies. We have also set out the main sources of uncertainty and directions for future work. Appendix~\ref{appendix-c-cross-domain-synthesis-tables} draws the three domains together, summarizing their harmonization status and separating the inputs that are externally anchored from those that remain priors awaiting elicitation. We hope this analysis proves useful for formulating common thresholds that meaningfully reduce AI risk.


\section*{Contributions}
W.S.A., M.B. and L.F.L. were the lead authors of the Cyber Risk, Biorisk, and Automated AI R\&D sections, respectively. 
M.G. coordinated and supervised the whole project. 
The initial idea of the project was proposed by Charbel-Raphael Segerie.
The work was done within the Supervised Program for Alignment Research (SPAR).
\clearpage
\bibliographystyle{plainnat}
\bibliography{referencesClaude}

@misc{alkarmi2026,
      title={Estimating the Social Cost of Corporate Data Breaches}, 
      author={Lina Alkarmi and Armin Sarabi and Mingyan Liu},
      year={2026},
      eprint={2603.21270},
      archivePrefix={arXiv},
      primaryClass={cs.CR},
      url={https://arxiv.org/abs/2603.21270}, 
}

@techreport{anthropic2025opus4card,
  author       = {{Anthropic}},
  title        = {Claude Opus 4 System Card},
  institution  = {Anthropic},
  year         = {2025},
  month        = may,
  type         = {Technical report},
}

@misc{anthropic2026a,
  author       = {{Anthropic}},
  title        = {Responsible Scaling Policy, Version 3.0},
  year         = {2026},
  institution  = {Anthropic},
  url          = {https://www.anthropic.com/responsible-scaling-policy/rsp-v3-0},
}

@techreport{anthropic2026b,
  author       = {{Anthropic}},
  title        = {Claude Mythos Preview System Card},
  institution  = {Anthropic},
  year         = {2026},
  month        = apr,
  type         = {Technical report},
}

@misc{anthropic2026c,
  author       = {{Anthropic}},
  title        = {Biorisk},
  year         = {2026},
  institution  = {Anthropic},
  url          = {https://red.anthropic.com/2025/biorisk/},
}

@misc{anthropic2026d,
  author       = {{Anthropic}},
  title        = {Responsible Scaling Policy, Version 3.1},
  year         = {2026},
  institution  = {Anthropic},
  url          = {https://www-cdn.anthropic.com/files/4zrzovbb/website/bf04581e4f329735fd90634f6a1962c13c0bd351.pdf},
}

@misc{anthropic2026e,
  author       = {{Anthropic}},
  title        = {Responsible Scaling Policy, Version 3.2},
  year         = {2026},
  institution  = {Anthropic},
  url          = {https://cdn.sanity.io/files/4zrzovbb/website/28c6241900d90410628a8a2003a5572faae4365a.pdf},
}

@misc{barrett2025,
  author       = {Barrett, S. and Murray, M. and Quarks, O. and Smith, M. and
                  Krys, J. and Campos, S. and Tlaie Boria, A. and Touzet, C. and
                  Hayrapet, S. and Heiding, F. and Nevo, O. and Swanda, A. and
                  Aguirre, J. and Gershovich, A. B. and Clay, E. and Fetterman, R. and
                  Fritz, M. and Juarez, M. and Mavroudis, V. and Papadatos, H.},
  title        = {Toward Quantitative Modeling of Cybersecurity Risks due to {AI} Misuse},
  year         = {2025},
  howpublished = {arXiv},
  eprint       = {2512.08864},
  archivePrefix= {arXiv},
  url          = {https://arxiv.org/abs/2512.08864},
}

@misc{sb53,
  author       = {{California Legislature}},
  title        = {{Senate Bill No. 53: Transparency in Frontier Artificial Intelligence Act}},
  year         = {2025},
  institution  = {{State of California}},
}

@misc{NYRaise,
  author       = {{New York State Legislature}},
  title        = {{Responsible AI Safety and Education Act (RAISE Act)}},
  year         = {2025},
  institution  = {{State of New York}},
}

@techreport{fbi2025,
  author       = {{Federal Bureau of Investigation Internet Crime Complaint Center}},
  title        = {Internet Crime Report 2024},
  institution  = {Federal Bureau of Investigation},
  year         = {2025},
  url          = {https://www.ic3.gov/AnnualReport/Reports/2024_IC3Report.pdf},
}

@misc{folkerts2026,
  author       = {Folkerts, L. and Payne, W. and Inman, S. and Giavridis, P. and
                  Skinner, J. and Deverett, S. and Aung, J. and Zorer, E. and
                  Schmatz, M. and Ghanem, M. and Wilkinson, J. and Steer, A. and
                  Hong, V. and Wang, J.},
  title        = {Measuring {AI} Agents' Progress on Multi-Step Cyber Attack Scenarios},
  year         = {2026},
  howpublished = {arXiv},
  eprint       = {2603.11214},
  archivePrefix= {arXiv},
  url          = {https://arxiv.org/abs/2603.11214},
}

@misc{googledeepmind2025,
  author       = {{Google DeepMind}},
  title        = {Frontier Safety Framework, Version 3.0},
  year         = {2025},
  month        = sep,
  institution  = {Google DeepMind},
  url          = {https://storage.googleapis.com/deepmind-media/DeepMind.com/Blog/strengthening-our-frontier-safety-framework/frontier-safety-framework_3.pdf},
}

@misc{gopal2025,
  author       = {Gopal, Anjali and Guest, Oliver and Besiroglu, Tamay and {Epoch AI}},
  title        = {{AI} and Biological Risk: Forecasting Key Capability Thresholds},
  year         = {2025},
  month        = oct,
  howpublished = {Epoch AI / EA Forum},
}

@misc{ho2025,
      title={A {Rosetta Stone} for {AI} Benchmarks}, 
      author={Anson Ho and Jean-Stanislas Denain and David Atanasov and Samuel Albanie and Rohin Shah},
      year={2025},
      eprint={2512.00193},
      archivePrefix={arXiv},
      primaryClass={cs.AI},
      url={https://arxiv.org/abs/2512.00193}, 
}

@techreport{ibm2025,
  author       = {{IBM Security}},
  title        = {Cost of a Data Breach Report 2025},
  institution  = {IBM},
  year         = {2025},
  url          = {https://www.ibm.com/reports/data-breach},
}

@misc{inagaki2023,
  author       = {Inagaki, T. and others},
  title        = {{LLMs} Can Generate Robotic Scripts from Goal-Oriented Instructions
                  in Biological Laboratory Automation},
  year         = {2023},
  howpublished = {arXiv},
  eprint       = {2304.10267},
  archivePrefix= {arXiv},
  url          = {https://arxiv.org/abs/2304.10267},
}

@misc{koessler2024,
  author       = {Koessler, L. and Schuett, J. and Anderljung, M.},
  title        = {Risk Thresholds for Frontier {AI}},
  year         = {2024},
  howpublished = {arXiv},
  eprint       = {2406.14713},
  archivePrefix= {arXiv},
  url          = {https://arxiv.org/abs/2406.14713},
}

@misc{lukosiute2026,
  author       = {Lukosiute, K. and Halstead, J. and Righetti, L.},
  title        = {Global Cybercrime Damages: A Baseline for Frontier {AI} Risk Assessment},
  year         = {2026},
  howpublished = {arXiv},
  eprint       = {2603.20570},
  archivePrefix= {arXiv},
  url          = {https://arxiv.org/abs/2603.20570},
}

@techreport{metr2025,
  author       = {{METR}},
  title        = {Common Elements of Frontier {AI} Safety Policies (December 2025 Update)},
  institution  = {METR},
  year         = {2025},
  url          = {https://metr.org/blog/2025-12-09-common-elements-of-frontier-ai-safety-policies},
}

@techreport{mouton2024,
  author       = {Mouton, Christopher A. and Lucas, Caleb and Guest, Ella},
  title        = {The Operational Risks of {AI} in Large-Scale Biological Attacks: Results of a Red-Team Study},
  year         = {2024},
  institution  = {RAND Corporation},
  number       = {RR-A2977-2},
  url          = {https://www.rand.org/pubs/research_reports/RRA2977-2.html},
}

@misc{zhang2025genai,
  author       = {Zhang, Zaixi and Chakraborty, Souradip and Bedi, Amrit Singh and others},
  title        = {Generative {AI} for Biosciences: Emerging Threats and Roadmap to Biosecurity},
  year         = {2025},
  howpublished = {arXiv},
  eprint       = {2510.15975},
  archivePrefix= {arXiv},
  url          = {https://arxiv.org/abs/2510.15975},
}

@article{qu2024crisprgpt,
  author       = {Qu, Yuanhao and Huang, Kaixuan and Yin, Ming and others},
  title        = {{CRISPR-GPT} for Agentic Automation of Gene-editing Experiments},
  year         = {2025},
  journal      = {Nature Biomedical Engineering},
  doi          = {10.1038/s41551-025-01463-z},
  url          = {https://www.nature.com/articles/s41551-025-01463-z},
}

@misc{cong2025labos,
  author       = {Cong, Le and others},
  title        = {{LabOS}: The {AI-XR} Co-Scientist That Sees and Works With Humans},
  year         = {2025},
  howpublished = {arXiv},
  eprint       = {2510.14861},
  archivePrefix= {arXiv},
  url          = {https://arxiv.org/abs/2510.14861},
}

@article{swanson2025virtuallab,
  author       = {Swanson, Kyle and Wu, Wesley and Bulaong, Nash L. and others},
  title        = {The Virtual Lab of {AI} agents designs new {SARS-CoV-2} nanobodies},
  year         = {2025},
  journal      = {Nature},
  doi          = {10.1038/s41586-025-09442-9},
  url          = {https://www.nature.com/articles/s41586-025-09442-9},
}

@misc{nvd2026cve4747,
  author       = {{NIST National Vulnerability Database}},
  title        = {{CVE-2026-4747}},
  year         = {2026},
  howpublished = {NVD},
  url          = {https://nvd.nist.gov/vuln/detail/CVE-2026-4747},
  note         = {FreeBSD RPCSEC\_GSS stack-overflow remote code execution; published 2026-03-26},
}

@misc{brent2025,
  author       = {Brent, Roger and McKelvey, Jr., T. Greg},
  title        = {Contemporary {AI} Foundation Models Increase Biological Weapons Risk},
  year         = {2025},
  howpublished = {arXiv},
  eprint       = {2506.13798},
  archivePrefix= {arXiv},
  url          = {https://arxiv.org/abs/2506.13798},
}

@misc{murray2025,
  author       = {Murray, Malcolm and Barrett, Steve and Papadatos, Henry and
                  Quarks, Otter and Smith, Matt and Tlaie Boria, Alejandro and
                  Touzet, Chlo\'e and Campos, Sim\'eon},
  title        = {A Methodology for Quantitative {AI} Risk Modeling},
  year         = {2025},
  institution  = {SaferAI},
  howpublished = {arXiv},
  eprint       = {2512.08844},
  archivePrefix= {arXiv},
  url          = {https://arxiv.org/abs/2512.08844},
}

@book{nasem2018,
  author       = {{National Academies of Sciences, Engineering, and Medicine}},
  title        = {Biodefense in the Age of Synthetic Biology},
  publisher    = {The National Academies Press},
  year         = {2018},
}

@misc{openai2025pf,
  author       = {{OpenAI}},
  title        = {Preparedness Framework, Version 2},
  year         = {2025},
  institution  = {OpenAI},
  url          = {https://cdn.openai.com/pdf/18a02b5d-6b67-4cec-ab64-68cdfbddebcd/preparedness-framework-v2.pdf},
}

@techreport{openai2025gpt5card,
  author       = {{OpenAI}},
  title        = {{GPT-5} System Card},
  institution  = {OpenAI},
  year         = {2025},
  month        = dec,
  type         = {Technical report},
}

@techreport{openai2026gpt55card,
  author       = {{OpenAI}},
  title        = {{GPT-5.5} System Card},
  institution  = {OpenAI},
  year         = {2026},
  month        = apr,
  type         = {Technical report},
}

@misc{righetti2025,
  author       = {Righetti, L.},
  title        = {Dual-Use {AI} Capabilities and the Risk of Bioterrorism:
                  Converting Capability Evaluations to Risk Assessments},
  year         = {2025},
  institution  = {GovAI},
}

@misc{sandbrink2023,
  author       = {Sandbrink, J.},
  title        = {Artificial Intelligence and Biological Misuse: Differentiating
                  Risks of Language Models and Biological Design Tools},
  year         = {2023},
  howpublished = {arXiv},
  eprint       = {2306.13952},
  archivePrefix= {arXiv},
  url          = {https://arxiv.org/abs/2306.13952},
}

@misc{securebio2024,
  author       = {{SecureBio}},
  title        = {Virology Capabilities Test (VCT)},
  year         = {2024},
  note         = {Published methodology and multi-select variant},
}

@misc{ukgov2024,
  author       = {{UK Government}},
  title        = {Frontier {AI} Safety Commitments, {AI} Seoul Summit 2024},
  year         = {2024},
}

@misc{usdot2026,
  author       = {{U.S. Department of Transportation}},
  title        = {Departmental Guidance on Valuation of a Statistical Life in Economic Analysis},
  year         = {2026},
  institution  = {U.S. Department of Transportation},
  url          = {https://www.transportation.gov/office-policy/transportation-policy/revised-departmental-guidance-on-valuation-of-a-statistical-life-in-economic-analysis},
}

@misc{ziosi2025,
  author       = {Ziosi and others},
  title        = {Safety Frameworks and Standards: A Comparative Analysis to
                  Advance Risk Management of Frontier {AI}},
  year         = {2025},
  month        = oct,
  howpublished = {Research Memo},
}

@misc{zhang2024,
  author       = {Zhang, Andy K. and others},
  title        = {Cybench: A Framework for Evaluating Cybersecurity Capabilities and Risks of Language Models},
  year         = {2024},
  howpublished = {arXiv},
  eprint       = {2408.08926},
  archivePrefix= {arXiv},
  url          = {https://arxiv.org/abs/2408.08926},
}

@misc{zhang2025,
  author       = {Zhang and others},
  title        = {BountyBench},
  year         = {2025},
}

@misc{wang2025cybergym,
  author       = {Wang, Zhun and Song, Dawn and others},
  title        = {CyberGym: Evaluating {AI} Agents' Real-World Cybersecurity Capabilities at Scale},
  year         = {2025},
  howpublished = {arXiv},
  eprint       = {2506.02548},
  archivePrefix= {arXiv},
  url          = {https://arxiv.org/abs/2506.02548},
}

@misc{zhang2026titanca,
  author       = {Zhang and others},
  title        = {{TitanCA}: Lessons from Orchestrating {LLM} Agents to Discover 100+ {CVE}s},
  year         = {2026},
  howpublished = {arXiv},
  eprint       = {2604.17860},
  archivePrefix= {arXiv},
  url          = {https://arxiv.org/abs/2604.17860},
}

@misc{chauvin2026epoch,
  author       = {Chauvin, Anson and Barry, Josh and Denain, Jean-Stanislas and Ho, Anson},
  title        = {Are {Mythos}' Cyber Capabilities Overhyped?},
  year         = {2026},
  howpublished = {Epoch AI, Gradient Updates},
  url          = {https://epoch.ai/gradient-updates/are-mythos-cyber-capabilities-overhyped},
}

\clearpage
\onecolumn
\appendix

\section{Threshold Equivalence Derivation}\label{appendix-a-threshold-equivalence-derivation}

This appendix provides the mathematical derivation supporting the threshold-equivalence claim in Section~\ref{sec:automated-ai-rd}. Under the assumption that ``several months'' in OpenAI's formulation means three months, the Anthropic and OpenAI automated AI R\&D thresholds are equivalent in terms of cumulative additional progress above trend. The derivation is given under both an exponential growth model \eqref{eq:appendix-exponential-growth} and a linear growth model \eqref{eq:appendix-linear-growth}; the equivalence result holds under both.

\subsection{Exponential Growth Model}\label{a.1-exponential-growth-model}

Assume that AI progress increases exponentially at rate $r$:

\begin{equation}\label{eq:appendix-exponential-growth}
C_r(t_0 + \Delta) = C_0 e^{r\Delta},
\end{equation}

where $C_r(t)$ is capability at time $t$ with rate of progress $r$, and $C_0$ is capability at time $t_0$. OpenAI's threshold corresponds to $r$ being five times $r_{2024}$ (the rate of progress in 2024), since:

\(C_{r}(t_{0} + \Delta) = C_{0}e^{(5r_{2024})\Delta} = C_{0}e^{r_{2024}(5\Delta)},\)

(a rate $5r_{2024}$ during a time $\Delta$ leads to the same progress as a rate $r_{2024}$ during a time $5\Delta$), while Anthropic's is just $2r_{2018\text{-}2024}$. However, Anthropic requires this to be maintained, on average, for 1 year, while OpenAI requires it for ``several months''. These two thresholds could be equivalent if we focus on additional progress with respect to baseline (and assume $r_{2024} = r_{2018\text{-}2024} \equiv r_0$). Let additional progress be defined as:

\(C_{r}(t_{0} + \Delta)/C_{r_{0}}(t_{0} + \Delta)\  = \ e^{(r - r_{0})\Delta}\),

where the equality follows from \eqref{eq:appendix-exponential-growth}. For Anthropic, the additional progress after one year is \(e^{(2r_{0} - r_{0})1} = e^{r_{0}}\), while for OpenAI, after $m$ months it is \(e^{(5r_{0} - r_{0})m/12} = e^{r_{0}m/3}\). Equating the two shows that the rate proposed by OpenAI, maintained for 3 months, yields the same additional progress as the rate proposed by Anthropic, maintained for 1 year.

\subsection{Linear Growth Model}\label{a.2-linear-growth-model}

If, instead, we assume linear increase at rate $r$:

\begin{equation}\label{eq:appendix-linear-growth}
C_r(t_0 + \Delta) = C_0 + r\Delta,
\end{equation}

OpenAI's threshold corresponds to $r$ being multiplied by 5, while Anthropic's corresponds to $r$ being multiplied by 2. The additional advance for Anthropic's threshold is $(2r_{2018\text{-}2024} - r_{2018\text{-}2024}) \cdot 1 = r_{2018\text{-}2024}$, while for OpenAI it is $(5r_{2024} - r_{2024}) \cdot m/12 = 4r_{2024}\,m/12$. Again, assuming $r_{2024} = r_{2018\text{-}2024}$, the additional advance is equal if $m=3$.

Under both growth models, the Anthropic and OpenAI thresholds imply equal cumulative additional capability gain when $m = 3$ months. Because OpenAI's formulation requires the elevated rate to be sustained for ``several months'' (interpreted here as three months) and because it is triggered by more advanced capability within a shorter window, it serves as the minimum common denominator adopted in the harmonized formulation proposed in Section~\ref{sec:automated-ai-rd}. The GDM threshold is compatible with this common denominator since it leaves open the precise meaning of AI progress ``substantially accelerating.''

\section{Cyber Risk Calibration Model}\label{appendix-b-cyber-risk-calibration-model}

This appendix contains the full quantitative calibration for the cyber risk model in Section~\ref{cyber-risk-and-expected-harm}. All numerical values are author-calibrated priors intended to illustrate the methodology. They are not derived through the IDEA expert-elicitation protocol that a mature application would require and should not be used as inputs to release decisions without independent validation.

\subsection{Baseline Cyber Harm Across AT1--AT5 Pathways}\label{b.1-baseline-cyber-harm-across-at1-at5-pathways}

Table~\ref{tab:b1-baseline} decomposes the USD~500 billion global cybercrime baseline into ten illustrative pathways across AT1--AT5. AT1--AT5 denotes attacker-tier sophistication, from low-skill, high-volume abuse to strategic national-scale compromise. The values are calibrated pathway parameters rather than independently verified raw event counts. Their purpose is to expose the model structure and identify where historical evidence is most needed.

\begin{longtable}[]{@{}lL{5.0cm}rrrr@{}}
\caption{Baseline cyber harm allocation across AT1--AT5 pathways.}\label{tab:b1-baseline}\\
\toprule
$AT$ & \textbf{Pathway} & $N_0$ & $P_0$ & \textbf{H per success} & \textbf{Baseline harm}\\
\midrule
\endfirsthead
\toprule
$AT$ & \textbf{Pathway} & $N_0$ & $P_0$ & \textbf{H per success} & \textbf{Baseline harm}\\
\midrule
\endhead
AT1 & Credential phishing / account takeover & 80M & 0.10 & USD 10K & USD 80B\\
AT1 & Commodity malware / low-skill intrusion & 20M & 0.10 & USD 20K & USD 40B\\
AT2 & BEC / social engineering & 2M & 0.17 & USD 250K & USD 85B\\
AT2 & Affiliate ransomware against SMEs & 500K & 0.20 & USD 550K & USD 55B\\
AT3 & Enterprise ransomware / domain compromise & 50K & 0.50 & USD 4M & USD 100B\\
AT3 & Enterprise data theft / extortion & 25K & 0.50 & USD 4M & USD 50B\\
AT4 & Zero-day chains against hardened organizations & 2.5K & 0.20 & USD 50M & USD 25B\\
AT4 & Supply-chain or SaaS compromise & 700 & 0.10 & USD 500M & USD 35B\\
AT5 & Critical infrastructure disruption & 60 & 0.10 & USD 3B & USD 18B\\
AT5 & Strategic national-scale compromise & 12 & 0.10 & USD 10B & USD 12B\\
 & \textbf{Total} & & & & \textbf{USD 500B}\\
\bottomrule
\multicolumn{6}{p{0.96\linewidth}}{\footnotesize\textit{Note.} $N_0$ and $P_0$ are calibrated baseline parameters. The table distributes the aggregate baseline across actor/pathway categories for transparent sensitivity analysis. Pathway allocations should be replaced by victimization survey data, breach telemetry, cyber insurance claims, and ransomware payment records.}
\end{longtable}

From Table~\ref{tab:b1-baseline} we recover the empirically estimated expected harm:

\begin{equation}\label{eq:appendix-baseline-harm}
\begin{aligned}
E[H_{\mathrm{baseline}}] &= 80\mathrm{B} + 40\mathrm{B} + 85\mathrm{B} + 55\mathrm{B} + 100\mathrm{B} \\
&\quad + 50\mathrm{B} + 25\mathrm{B} + 35\mathrm{B} + 18\mathrm{B} + 12\mathrm{B} \\
&= 500\ \mathrm{B\ USD/year}
\end{aligned}
\end{equation}

This allocation is illustrative and back-fit to the USD 500 billion anchor: the ten rows are constructed to sum to that total, so the per-pathway $N_0$, $P_0$, and $h$ are not independent priors, and changing one row requires another to absorb the difference if the anchor is held fixed. The AT1--AT2 dominance is therefore a modelling choice rather than an independent finding. A mature version should replace the whole partition with independent per-pathway anchors, drawn from victimization surveys, breach telemetry, ransomware-payment records, and cyber-insurance claims, so that each row carries its own evidence and can be rejected on its own terms.

\subsection{Estimating AI-Enabled Uplift}\label{b.2-estimating-ai-enabled-uplift}

The following calculation illustrates the methodology for a model at Mythos Preview capability level. Mythos Preview is the reference model because it is the most capable publicly evaluated model without an active Anthropic policy threshold, making the release-condition question directly actionable: which release conditions, if any, bring a model at this capability level below the USD~1B statutory benchmark?

Throughout the uplift table, $P_{\mathrm{AI},j}$ denotes the success rate of the marginal, AI-enabled attempts on pathway $j$, not the population success rate across all attackers on it. That distinction keeps the small AT4--AT5 entries coherent: AI draws new, lower-skill actors toward elite intrusion and most of them fail, so the marginal success rate on AT4--AT5 sits below the incumbent baseline $P_0$, even as AI raises the success rate of the lower tiers it genuinely assists.

The primary TLO figure used here is 6/10 full-chain completions, drawn from the most recent AISI evaluation of Mythos Preview. An earlier evaluation reported 3/10; the 6/10 figure is used as the more recent estimate. The proposed minimum floor, non-zero full-chain completion, is unaffected by which figure is used. For AT4--AT5, $P_{\mathrm{AI}}$ values are kept small but non-zero in the central calibration because Mythos-level capability is not assumed to materially uplift nation-state-tier operations requiring rare target access, long-term stealth, and strategic intent. The combined AT4--AT5 contribution is approximately USD~0.25B of the USD~67.90B total, so setting these values to zero would not change the headline result; the small non-zero values are retained for transparency.

A caveat on the volume assumption: this model treats additional AI-enabled volume as a fraction of baseline volume, represented by $q$. Under full autonomous operation, machine-speed attack pipelines rather than human-speed operations could allow a single actor to launch orders of magnitude more attempts than the human-population-constrained baseline implies. Attack volume may become a function of AI capability rather than actor population. The harm estimates here are therefore conservative with respect to volume.

\begin{longtable}[]{@{}lL{4.5cm}rL{2.2cm}r@{}}
\caption{AI-enabled uplift at Mythos Preview level, public API baseline.}\label{tab:b2-uplift}\\
\toprule
$AT$ & \textbf{Pathway} & $q$ & $P_{\mathrm{AI}}$ \textbf{(Mythos)} & \textbf{AI-enabled harm}\\
\midrule
\endfirsthead
\toprule
$AT$ & \textbf{Pathway} & $q$ & $P_{\mathrm{AI}}$ \textbf{(Mythos)} & \textbf{AI-enabled harm}\\
\midrule
\endhead
AT1 & Credential phishing / account takeover & 0.20 & 0.12 & USD 19.20B\\
AT1 & Commodity malware / low-skill intrusion & 0.15 & 0.12 & USD 7.20B\\
AT2 & BEC / social engineering & 0.15 & 0.20 & USD 15.00B\\
AT2 & Affiliate ransomware against SMEs & 0.10 & 0.30 & USD 8.25B\\
AT3 & Enterprise ransomware / domain compromise & 0.10 & 0.60 [TLO: 6/10] & USD 12.00B\\
AT3 & Enterprise data theft / extortion & 0.10 & 0.60 [TLO: 6/10] & USD 6.00B\\
AT4 & Zero-day chains against hardened orgs & 0.05 & 0.01 [zero-sensitivity recommended] & USD 0.06B\\
AT4 & Supply-chain or SaaS compromise & 0.05 & 0.01 & USD 0.18B\\
AT5 & Critical infrastructure disruption & 0.02 & 0.002 & USD 0.01B\\
AT5 & Strategic national-scale compromise & 0.02 & 0.001 & USD 0.00B\\
 & \textbf{Total} & & & \textbf{USD 67.90B}\\
\bottomrule
\multicolumn{5}{p{0.96\linewidth}}{\footnotesize\textit{Note.} Each cell is $q \times N_0 \times P_{\mathrm{AI}} \times H$, with $N_0$ and $H$ taken from Table~\ref{tab:b1-baseline}. On AT1--AT3 the modelled $P_{\mathrm{AI}}$ exceeds the baseline $P_0$, so the uplift combines additional volume with a success-rate gain rather than expanding volume alone. The headline figure is dominated by AT1--AT2 high-volume pathways (USD 49.65B), not by AT3 elite-intrusion capability (USD 18.00B). At GPT-5.5 level (2/10 TLO, AT3 $P_{\mathrm{AI}} = 0.20$), the total is approximately USD 55.90B. At pre-frontier baseline (0/10 TLO, no AT3 contribution), it is approximately USD 49.90B. The calibration should not be read as a claim that any nonzero AT3 capability alone generates the headline figure.}
\end{longtable}

\begin{longtable}[]{@{}L{2.8cm}L{2.4cm}L{2.4cm}L{2.5cm}L{2.8cm}@{}}
\caption{Sensitivity of additional AI-enabled harm to baseline assumption.}\label{tab:b3-sensitivity}\\
\toprule
\textbf{Release condition} & $H_0 = \text{USD}\ 100\text{B}$ & $H_0 = \text{USD}\ 500\text{B}$ & $H_0 = \text{USD}\ 1{,}000\text{B}$ & \textbf{Above USD 1B annual-harm reference?}\\
\midrule
\endfirsthead
\toprule
\textbf{Release condition} & $H_0 = \text{USD}\ 100\text{B}$ & $H_0 = \text{USD}\ 500\text{B}$ & $H_0 = \text{USD}\ 1{,}000\text{B}$ & \textbf{Above USD 1B annual-harm reference?}\\
\midrule
\endhead
Open weights & USD 27.2B & USD 135.8B & USD 271.6B & Yes (all)\\
Public API & USD 13.6B & USD 67.9B & USD 135.8B & Yes (all)\\
Safety-filtered API & USD 7.9B & USD 39.3B & USD 78.7B & Yes (all)\\
Trusted API & USD 0.06B & USD 0.32B & USD 0.63B & No (all)\\
Internal only & $<$USD 0.01B & $<$USD 0.01B & $<$USD 0.01B & No\\
\bottomrule
\multicolumn{5}{p{0.96\linewidth}}{\footnotesize\textit{Note.} Each row is the pathway-weighted aggregation of the release-condition scalars in Table~\ref{tab:b4-scalars} against the per-tier AI-enabled harm in Table~\ref{tab:b2-uplift}, so this table is the direct roll-up of that matrix rather than a flat per-condition discount. Values scale linearly with $H_0$. The Public API conclusion holds across the whole anchor range. The Trusted API figure now sits below the USD~1B annual-harm reference across the entire range, but it reaches USD~0.63B at the top of the anchor interval, within a factor of two of the reference, and it rests on the uncited exposure scalars of Table~\ref{tab:b4-scalars}; it should be read as an audit target, not a settled verdict.}
\end{longtable}

\subsection{Release Conditions as Exposure Scalars}\label{b.3-release-conditions-as-exposure-scalars}

Release conditions determine how much of the AI-enabled pathway risk remains accessible to relevant actors. Open weights are modeled as higher exposure than ordinary public API access because they permit unrestricted copying, fine-tuning, scaffolding, removal of safety layers, and integration into autonomous tooling. Trusted API is modeled as lower exposure, but not zero, because sophisticated AT3--AT5 actors may obtain access through front companies, compromised accounts, insiders, or indirect procurement. Internal-only use is modeled as residual exposure from insider misuse, compromise, or model theft.

The exposure scalar decomposes as:

\begin{equation}\label{eq:appendix-release-vector}
s_{r,j} = A_{r,j} \times C_{r,j} \times T_{r,j} \times D_{r,j} + L_{r,j}
\end{equation}

where $A$ is actor access, $C$ is capability retention, $T$ is throughput, $D$ is detection failure, and $L$ is leakage. The scalar is indexed by both release condition $r$ and pathway $j$. The Trusted API row is non-uniform across pathways: $A_{\mathrm{trusted},j}$ runs from 0.01 at AT1, because low-skill actors rarely pass KYC, to near 1.00 at AT5, because nation-state actors can pose as legitimate users. This produces a roughly 100$\times$ range within a single release condition.

\begin{longtable}[]{@{}L{2.6cm}ccccc@{}}
\caption{Release-condition exposure scalar matrix ($s_{r,j}$).}\label{tab:b4-scalars}\\
\toprule
\textbf{Release condition} & \textbf{AT1} & \textbf{AT2} & \textbf{AT3} & \textbf{AT4} & \textbf{AT5}\\
\midrule
\endfirsthead
\toprule
\textbf{Release condition} & \textbf{AT1} & \textbf{AT2} & \textbf{AT3} & \textbf{AT4} & \textbf{AT5}\\
\midrule
\endhead
Open weights & 2.00 & 2.00 & 2.00 & 2.00 & 2.00\\
Public API (ref.) & 1.00 & 1.00 & 1.00 & 1.00 & 1.00\\
Safety-filtered API & 0.60 & 0.58 & 0.55 & 0.50 & 0.45\\
Trusted API & 0.001 & 0.005 & 0.009 & 0.045 & 0.090\\
Internal only & $\sim$0 & $\sim$0 & $\sim$0 & $\sim$0 & $\sim$0\\
\bottomrule
\multicolumn{6}{p{0.96\linewidth}}{\footnotesize\textit{Note.} These values are uncited author priors and audit targets, not measured safeguard properties. Establishing them empirically would require measured inputs across several categories: KYC false-accept rates (for example against NIST SP 800-63A IAL2), presentation-attack detection, jailbreak success, account-compromise rates, and model-weight security. Sourcing these is deferred to dedicated open-source-intelligence work. The Trusted API row is non-uniform because sophisticated actors (AT4--AT5) are harder to exclude through access controls.}
\end{longtable}

\subsection{Required Safeguard Performance}\label{b.4-required-safeguard-performance-performance}

The safeguard question can be stated directly: how much public-access exposure must be removed before the model falls below a given acceptable-harm benchmark? The answer can be written as a maximum allowable exposure scalar:

\begin{equation}\label{eq:appendix-allowed-exposure}
s_{\max} = \frac{H_{\mathrm{benchmark}}}{E[H_{\mathrm{AI,public}}]}
\end{equation}

At Mythos Preview level, using the USD~67.90B central estimate:

\begin{equation}\label{eq:appendix-smax-one-billion}
s_{\max}(\text{USD}\ 1\text{B}) = \frac{1.0\text{B}}{67.90\text{B}} = 0.0147
\end{equation}

\begin{equation}\label{eq:appendix-smax-five-hundred-million}
s_{\max}(\text{USD}\ 500\text{M}) = \frac{0.5\text{B}}{67.90\text{B}} = 0.0074
\end{equation}

Safeguards must reduce effective exposure to roughly 1.5 percent of public API access to fall below the USD~1B SB-53 benchmark, and to roughly 0.7 percent for a USD~500M benchmark. These ratios scale linearly with the USD~500 billion baseline, whose 90 percent confidence interval spans an order of magnitude, so the honest band runs from a few tenths of a percent to a few percent; the three-figure values above (0.0147 and 0.0074) are central illustrations, not precise targets. These are audit targets; they do not prove that any existing trusted-access or safeguard regime achieves such reductions. Trusted access is therefore not a label but a measurable exclusion target. A company should provide independent evidence that KYC, monitoring, rate limits, jailbreak resistance, model-weight security, and account-compromise controls achieve the necessary exposure reduction.

\subsection{Kill Chain Decomposition}\label{b.5-kill-chain-decomposition}

Table~\ref{tab:b5-killchain} maps attack phases to pathways and provides per-phase conditional success probabilities. AT3 domain compromise values are from \citet{barrett2025}, Table~12; other columns are author-calibrated estimates. This table is a structural illustration of the AND-gate decomposition in \eqref{eq:cyber-kill-chain}: it shows how per-phase probabilities compose into a full-chain $P_{\mathrm{success},j}$. Its column products are not the values used in the headline harm totals. Those totals read the AT3 pathway success probability off the TLO evidence in Table~\ref{tab:tlo} ($P_{\mathrm{AI}} = 0.60$ at Mythos Preview level), whereas the AT3 column here multiplies out to 0.084. The two numbers are estimates of the same quantity, the full-chain completion rate, reached by different routes, and they differ by roughly sevenfold. We flag this as an open modelling question rather than paper over it, and we do not substitute the 0.084 product into the harm equations. Lateral movement and exfiltration are marked as TLO-relevant bottleneck phases, where AI uplift propagates multiplicatively through the product.

{\footnotesize
\begin{longtable}[]{@{}L{3.0cm}C{1.65cm}C{1.35cm}C{1.9cm}C{1.35cm}C{1.45cm}C{1.5cm}@{}}
\caption{Kill-chain phase matrix. Per-phase conditional success probability $p_{k,j}$. N/A = phase does not apply. Rows labeled ``TLO bottleneck'' indicate phases where AI uplift is especially consequential.}\label{tab:b5-killchain}\\
\toprule
\textbf{Phase} & \textbf{AT1:}\newline Phishing & \textbf{AT2:}\newline BEC & \textbf{AT2:}\newline Ransomware & \textbf{AT3:}\newline Domain & \textbf{AT4:}\newline Zero-day & \textbf{AT5:}\newline Strategic\\
\midrule
\endfirsthead
\toprule
\textbf{Phase} & \textbf{AT1:}\newline Phishing & \textbf{AT2:}\newline BEC & \textbf{AT2:}\newline Ransomware & \textbf{AT3:}\newline Domain & \textbf{AT4:}\newline Zero-day & \textbf{AT5:}\newline Strategic\\
\midrule
\endhead
Reconnaissance & 1.00 & 1.00 & 1.00 & 1.00 & 1.00 & 1.00\\
Initial access & 0.40 & 0.70 & 0.60 & 0.60 & 0.20 & 0.10\\
Execution & N/A & 0.60 & 0.50 & 0.50 & 0.70 & 0.80\\
Privilege escalation & N/A & N/A & 0.70 & 0.70 & 0.80 & 0.85\\
Lateral movement\newline(\emph{TLO bottleneck}) & N/A & N/A & 0.65 & 0.65 & 0.75 & 0.80\\
Collection & N/A & N/A & 0.90 & 0.90 & 0.85 & 0.90\\
Exfiltration\newline(\emph{TLO bottleneck}) & N/A & N/A & N/A & 0.85 & 0.80 & 0.85\\
Impact / monetization & 0.25 & 0.40 & 0.30 & 0.80 & 0.25 & 0.10\\
\midrule
$P_{\mathrm{success},j}$ & 0.100 & 0.168 & 0.037 & 0.084 & 0.014 & 0.004\\
\bottomrule
\multicolumn{7}{p{0.96\linewidth}}{\footnotesize\textit{Note.} AT3 domain compromise values are adapted from \citet{barrett2025}, Table~12 (OC3 SME Ransomware). Other values are author-calibrated estimates. ``TLO bottleneck'' phases are those emphasized by \citet{folkerts2026}: lateral movement (NTLM relay) and exfiltration.}
\end{longtable}
}

\section{Cross-Domain Synthesis Tables}\label{appendix-c-cross-domain-synthesis-tables}


Table~\ref{tab:domain-comparison} summarizes the harmonization status of each domain, and Table~\ref{tab:uncertainty} classifies the principal inputs and claims by the kind of evidence each would require.

{\footnotesize
\renewcommand{\arraystretch}{1.3}
\begin{longtable}[]{@{}L{1.6cm}L{2.3cm}L{2.5cm}L{3.0cm}L{3.1cm}L{3.1cm}@{}}
\caption{Cross-domain comparison of harmonization status.}\label{tab:domain-comparison}\\
\toprule
\textbf{Domain} & \textbf{Common primitive} & \textbf{Harmonization status} & \textbf{Main blocker} & \textbf{Paper contribution} & \textbf{Next empirical step}\\
\midrule
\endfirsthead
\toprule
\textbf{Domain} & \textbf{Common primitive} & \textbf{Harmonization status} & \textbf{Main blocker} & \textbf{Paper contribution} & \textbf{Next empirical step}\\
\midrule
\endhead
Cyber & Expected annual harm (USD) & Tractable; preliminary floor proposed & Calibration uncertainty; lack of release-condition audit data & Expected-harm translation; non-zero TLO completion floor; release-condition decomposition & Historically calibrated pathway estimates and audited safeguard performance\\
Biorisk & Expected casualties & Methodology applies; no quantitative floor yet & No unsaturated, validated, full-chain KRI; near-zero base rate & Diagnosis of the missing evidence; staged kill-chain floor structure & Delphi expert elicitation of per-stage uplift; Monte Carlo aggregation\\
Automated AI R\&D & Rate of capability progress (ECI) & Achievable now; operational floor proposed & Attribution to AI automation; benchmark and harness drift & Quantitative floor (5$\times$ trend for 3 months) covering the three main frontier companies & Industry-wide baseline; attribution evidence\\
\bottomrule
\multicolumn{6}{p{0.96\linewidth}}{\footnotesize\textit{Note.} The cyber and biorisk floors take expected harm as the common primitive following \citet{koessler2024,murray2025}; the automated AI R\&D floor uses a rate-of-progress primitive because no single harm pathway applies.}
\end{longtable}
}

{\footnotesize
\renewcommand{\arraystretch}{1.3}
\begin{longtable}[]{@{}L{3.4cm}L{2.8cm}L{4.2cm}L{2.2cm}L{3.4cm}@{}}
\caption{Uncertainty classification: structural assumptions versus empirical placeholders.}\label{tab:uncertainty}\\
\toprule
\textbf{Parameter or claim} & \textbf{Current status} & \textbf{Evidence needed} & \textbf{Importance} & \textbf{Likely owner}\\
\midrule
\endfirsthead
\toprule
\textbf{Parameter or claim} & \textbf{Current status} & \textbf{Evidence needed} & \textbf{Importance} & \textbf{Likely owner}\\
\midrule
\endhead
Cyber global baseline harm & Externally anchored estimate & Reconciliation of damage estimates with reported-loss data & High (scales all results) & Cyber-economics researchers\\
Cyber pathway allocation & Author-calibrated prior & Victimization surveys, breach telemetry, ransomware records & Medium (sensitivity-bounded) & Incident-data holders\\
Cyber release-condition scalars & Author-calibrated prior & KYC, monitoring, jailbreak, and weight-security audits & High (decides release verdict) & Independent safeguard auditors\\
TLO as cyber KRI & Proposed primary KRI & Continued longitudinal, multi-company cyber-range evaluation & High (defines the trigger) & Cyber-eval bodies (e.g.\ AISI)\\
AT4--AT5 uplift assumptions & Author-calibrated prior (small, non-zero) & Analysis of AI uplift to nation-state-tier operations & Low (small harm share) & Threat-intelligence analysts\\
Biorisk: no full-chain validated KRI & Stated claim & Audit of public bio-evaluations against the three criteria & High (blocks the floor) & Biosecurity evaluation community\\
Biorisk base-rate problem & Stated limitation & Structured counterfactual elicitation design & High (frames all estimates) & Biosecurity and biodefense experts\\
Biorisk expert elicitation design & Proposed, not completed & Delphi panel with calibrated, diverse experts & High (the missing input) & Bio elicitation consortium\\
AI R\&D ECI metric & Proposed primary metric & Robustness to benchmark and harness changes & Medium (alternatives exist) & Forecasting and eval reviewers\\
AI R\&D attribution to automation & Open question & Trend-break evidence in AI R\&D inputs and benchmarks & Medium (trigger vs.\ audit) & AI progress researchers\\
\bottomrule
\multicolumn{5}{p{0.96\linewidth}}{\footnotesize\textit{Note.} ``Likely owner'' denotes the kind of party best positioned to supply or contest the evidence, not an endorsement by any named organization. Importance is judged relative to whether the input could change a threshold-crossing verdict.}
\end{longtable}
}

\end{document}